\documentclass[10pt,journal,compsoc]{IEEEtran}
\ifCLASSOPTIONcompsoc
  \usepackage[nocompress]{cite}
\else
  \usepackage{cite}
\fi
\usepackage{times}
\usepackage{epsfig}
\usepackage{graphicx}
\usepackage{amsmath}
\usepackage{amssymb}

\usepackage{ifthen}
\usepackage{color}
\usepackage{bigstrut}
\usepackage{multirow}
\usepackage{multicol}
\usepackage{mathtools}
\usepackage[normalem]{ulem}
\usepackage{algorithm}
\usepackage{colortbl}
\usepackage{algpseudocode}
\usepackage[pagebackref=true,breaklinks=true,colorlinks,citecolor=blue,linkcolor=blue,bookmarks=false]{hyperref}

\DeclarePairedDelimiter\abs{\lvert}{\rvert}%

\newcommand{\etal}{\textit{et al}. }
\definecolor{frenchblue}{rgb}{0.0, 0.45, 0.73}
\definecolor{gray}{rgb}{0.5,0.5,0.5} 
\definecolor{green}{rgb}{0, 0.4, 0} 
\definecolor{orange}{rgb}{1, 0.5, 0} 	
\definecolor{mahogany}{rgb}{0.75, 0.25, 0.0}
\definecolor{purple}{rgb}{0.6, 0, 0.6}
\definecolor{darkgreen}{rgb}{0, 0.4, 0.4} 
\definecolor{teal}{rgb}{0.0, 0.5, 0.5}
\definecolor{aaaa}{rgb}{0.55, 0.1, 0.7}
\definecolor{red}{rgb}{1.0, 0, 0}

\newboolean{revising}
\setboolean{revising}{false}
\ifthenelse{\boolean{revising}}{

	\newcommand{\revise}[1]{\textcolor{blue}{#1}}
}{

	\newcommand{\revise}[1]{#1}
}

\begin{document}
\title{BiFuse++: Self-supervised and Efficient Bi-projection Fusion for 360{\boldmath$^\circ$} Depth Estimation}
\author{
	Fu-En Wang,
	Yu-Hsuan Yeh,
	Yi-Hsuan Tsai,
	Wei-Chen Chiu,
	and~Min Sun
	\IEEEcompsocitemizethanks{
		\IEEEcompsocthanksitem F.-E. Wang and M. Sun are with the Department
		of Electrical Engineering, National Tsing Hua University, Hsinchu, Taiwan.\protect\\
		E-mail: fulton84717@gapp.nthu.edu.tw
		\IEEEcompsocthanksitem Y.-H. Tsai is with Phiar Technologies, USA.
		\IEEEcompsocthanksitem Y.-H. Yeh and W.-C. Chiu are with the Department of Computer Science, National Yang Ming Chiao Tung University, Hsinchu, Taiwan.
            \IEEEcompsocthanksitem Source code is available on \url{https://github.com/fuenwang/BiFusev2}
	}
	\thanks{© 2022 IEEE.  Personal use of this material is permitted.  Permission from IEEE must be obtained for all other uses, in any current or future media, including reprinting/republishing this material for advertising or promotional purposes, creating new collective works, for resale or redistribution to servers or lists, or reuse of any copyrighted component of this work in other works.}
}

\markboth{}%
{}

\IEEEtitleabstractindextext{%
\begin{abstract}
    Due to the rise of spherical cameras, monocular 360$^\circ$ depth estimation becomes an important technique for many applications (e.g., autonomous systems). Thus, state-of-the-art frameworks for monocular 360$^\circ$ depth estimation such as bi-projection fusion in BiFuse are proposed. To train such a framework, a large number of panoramas along with the corresponding depth ground truths captured by laser sensors are required, which highly increases the cost of data collection. Moreover, since such a data collection procedure is time-consuming, the scalability of extending these methods to different scenes becomes a challenge.
    To this end, self-training a network for monocular depth estimation from 360$^\circ$ videos is one way to alleviate this issue.
    However, there are no existing frameworks that incorporate bi-projection fusion into the self-training scheme, which highly limits the self-supervised performance since bi-projection fusion can leverage information from different projection types. In this paper, we propose BiFuse++ to explore the combination of bi-projection fusion and the self-training scenario. To be specific, we propose a new fusion module and Contrast-Aware Photometric Loss to improve the performance of BiFuse and increase the stability of self-training on real-world videos. We conduct both supervised and self-supervised experiments on benchmark datasets and achieve state-of-the-art performance.
    
\end{abstract}

}

\maketitle

\IEEEdisplaynontitleabstractindextext

%
\IEEEpeerreviewmaketitle

\IEEEraisesectionheading{\section{Introduction}\label{sec:introduction}}
\begin{figure*}[t]
    \centering
    \includegraphics[width=\textwidth]{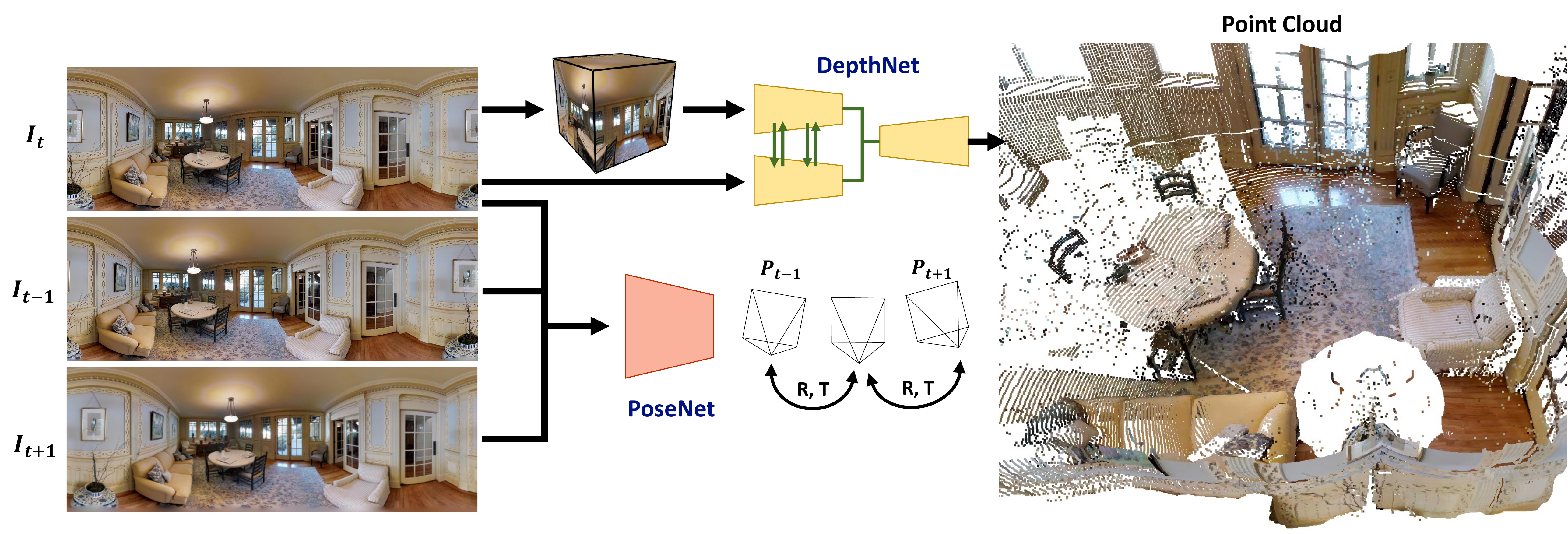}
    \caption{Our BiFuse++ is a self-supervised framework of monocular 360$^\circ$ depth estimation. The depth estimation network (DepthNet) is a bi-projection architecture consisting of two encoders and a shared decoder. The inputs of DepthNet are equirectangular and cubemap projections of reference panorama $I_t$. Between each adjacent layer of encoders, the feature maps of two projections are fused by our proposed fusion module (green arrows). To achieve self-supervised learning, an additional network (PoseNet) takes three adjacent panoramas ($I_{t-1}$, $I_t$, and $I_{t+1}$) in a video sequence as inputs and infers the corresponding camera motions ($P_{t-1}$ and $P_{t+1}$). We then compute the photo consistency error based on the predicted depth map and camera motions to jointly train the two networks.}
    \label{fig:teaser}
\end{figure*}
\IEEEPARstart{D}{epth} estimation from a single image is a crucial technique for many applications. For indoor autonomous systems, the geometric information of depth maps is necessary to improve navigation and exploration efficiency. Moreover, a thoughtful understanding of the environment provided by depth maps is also required to ensure the safety of the surrounding people. Traditionally, depth maps are captured by scanners such as LiDARs, structured light, or other time-of-flight (ToF) sensors. Such sensors are typically costly and, thus, have limited usages depending on the scenarios. With the advance of deep neural networks, depth estimation from the perspective cameras becomes a possible solution for these tasks. For instance, FCRN~\cite{FCRN} is one popular framework for monocular depth estimation. However, most existing frameworks are designed only for perspective cameras with limited field-of-view (FoV), whereas some depth sensors (e.g., LiDARs) offer 360$^\circ$ field-of-view.

With the rising availability of consumer-level spherical cameras, the omnidirectional camera turns into a good choice for indoor autonomous systems. By capturing the 360$^\circ$ information into a single panorama, 360$^\circ$ cameras can significantly increase the navigation and exploration efficiency. Coming along with this, 360$^\circ$ perception has become an important topic in computer vision. For instance, OmniDepth~\cite{omnidepth} and BiFuse~\cite{bifuse} are frameworks for monocular 360$^\circ$ depth estimation. These frameworks use the ground truth depth maps captured by depth sensors as the supervisory signals to train the network. In general, there are two major issues for such 360$^\circ$ depth estimation frameworks: \revise{1) Unlike common perspective cameras, the distortion introduced by equirectangular projection is extremely large, especially near the north and south poles on equirectangular coordinates. 2) the large number of depth maps captured by sensors is necessary to train the networks and thus highly increases the cost of data collection. Since the distortion can be removed by converting a single equirectangular image into several perspective ones where each of them only covers limited FoVs (e.g., cubemap projection), BiFuse~\cite{bifuse} utilizes the combination of equirectangular and cubemap projection as input to the framework. In this way, the 360$^\circ$ context can be preserved with the equirectangular projection, while the areas with large distortions can be resolved by the cubemap one. However, it is still necessary for the training of BiFuse to adopt a large-scale dataset consisting of precise depth ground truths from depth sensors, and the cost of data collection is usually large since the depth sensors with 360$^\circ$ FoV are expensive and not affordable by most consumers. Therefore, the requirement for a large amount of ground truth depth maps still prevents BiFuse from being extended to various scenes, thus reducing its scalability.}


To reduce the cost of data collection, self-supervised learning approaches of monocular depth estimation like SfMLearner~\cite{sfmlearner} are designed for normal FoV cameras. By utilizing SfMLearner and the cubemap projection, 360-SelfNet~\cite{ouraccv} is the first framework for self-supervised 360$^\circ$ depth estimation. However, only using cubemap and cubepadding~\cite{cubepadding} cannot provide complete information and also harms the stability of depth consistency around the cubemap boundary as addressed in \cite{bifuse}. To this end, combining both the equirectangular and cubemap projections appear to be a potential solution for self-supervised 360$^\circ$ depth estimation, in which such a combination has not been studied in this field.



In this paper, we extend the previous BiFuse~\cite{bifuse} work and propose an advanced framework ``BiFuse++'' for self-supervised monocular 360$^\circ$ depth estimation.
As illustrated in Figure~\ref{fig:teaser}, there are two networks, DepthNet and PoseNet, in our framework. DepthNet first estimates the depth map of the reference panorama $I_t$ and PoseNet estimates the corresponding camera motions between adjacent panoramas ($I_{t-1}$ and $I_{t+1}$). We then compute their photo consistency error to achieve self-supervised training. Our DepthNet is a bi-projection architecture which takes equirectangular and cubemap projections as inputs ($I_t$) to estimate the corresponding depth map. Motivated by UniFuse~\cite{unifuse}, we adopt a single decoder to unify feature maps from equirectangular and cubemap branches. To improve the performance and efficiency, we propose a new fusion module to exchange the information between different projections. Unlike UniFuse that simply infers equirectangular and cubemap feature maps from two independent encoders, our fusion module first fuses feature maps from two projections, and then the fused ones are further passed into the next layers of encoders. In this way, our encoders can directly retrieve the information from another branch and preserve more complete details on the predicted depth maps.

\begin{figure}[t]
    \centering
    \includegraphics[width=\columnwidth]{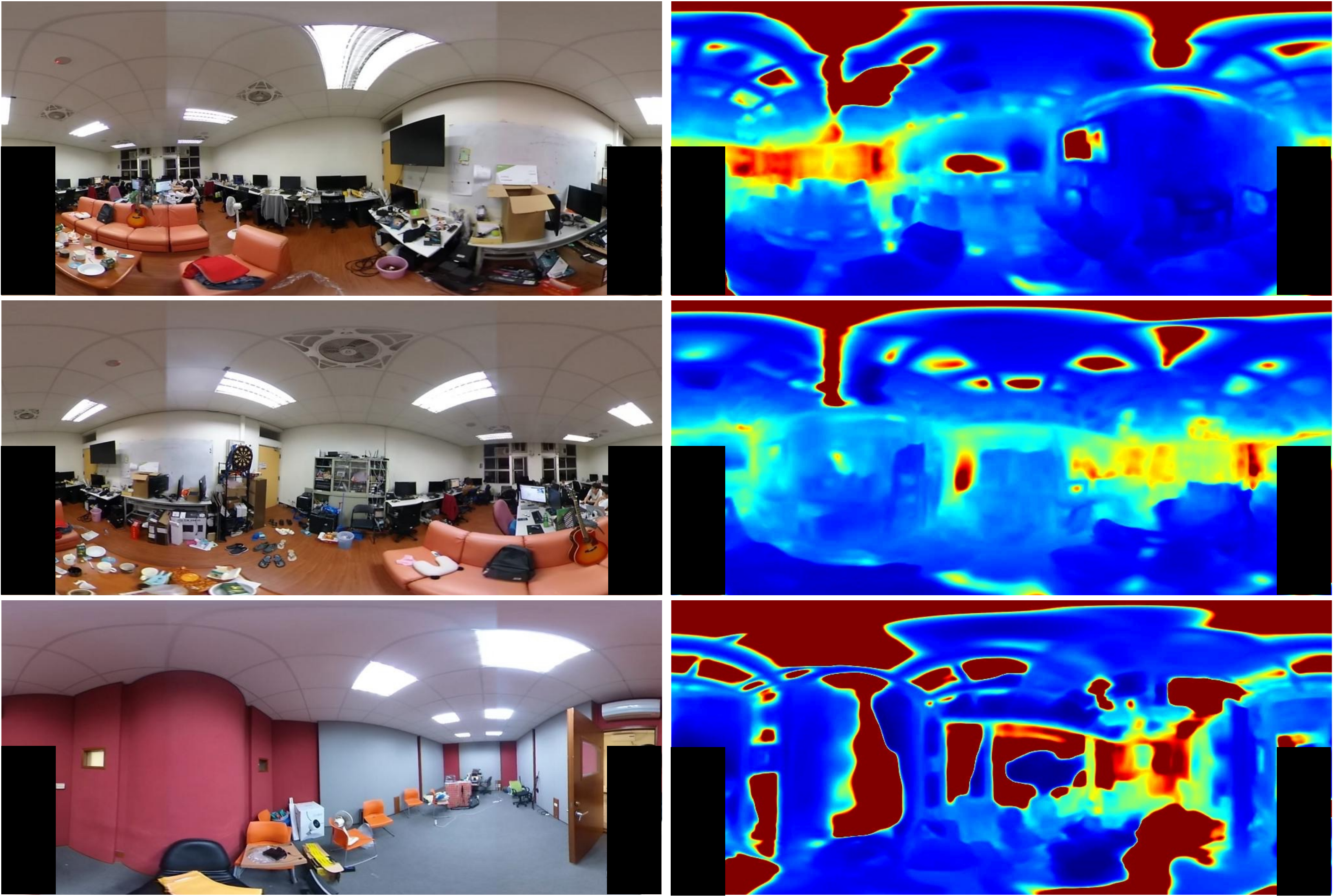}
    \caption{360-SelfNet~\cite{ouraccv} trained on real-world images. The spherical photometric loss cannot deal with the low-texture area and thus produces unstable depth maps (red indicates a large depth value). Note that we mask out the photographer at the bottom left and right region.}
    \label{fig:accv-realworld}
\end{figure}

\revise{To infer the camera motions between adjacent panoramas, 360-SelfNet~\cite{ouraccv} adopts an additional network that takes the concatenation of adjacent panoramas as input and uses an encoder to extract the camera motions. In addition, a decoder consisting of transposed convolutional layers is adopted to infer the occlusion masks that are further leveraged in photometric loss. Instead of following 360-SelfNet to adopt an encoder-decoder architecture, our PoseNet is a single encoder that takes three adjacent panoramas ($I_{t-1}$, $I_t$, and $I_{t+1}$) as input, and infers the backward and forward camera motions ($P_{t-1}$ and $P_{t+1}$), i.e., the camera motions from $I_t$ to $I_{t-1}$ and from $I_t$ to $I_{t+1}$. We then directly adopt additional convolutional layers in the encoder to extract occlusion masks at different scales.}
With the predicted depth map and camera motions, we can achieve self-supervised training based on the photo consistency assumption. However, we find that the spherical photometric loss proposed by 360-SelfNet has a degeneration problem in low-texture areas and severely harms the training performance in real-world videos (see Figure~\ref{fig:accv-realworld}). To this end, we propose ``Contrast-Aware Photometric Loss (CAPL)'' to deal with the degeneration. In addition to achieving self-supervised training on 360$^\circ$ videos, our BiFuse++ is also efficient and effective to be adopted in supervised training.

%
%

To validate the applicability of BiFuse++, we conduct extensive experiments under both supervised and self-supervised scenarios. For the self-supervised scenario, we perform experiments on the PanoSUNCG~\cite{ouraccv} dataset. For the supervised scenario, we evaluate our method on Matterport3D~\cite{mp3d}, Stanford2D3D~\cite{stfd}, and PanoSUNCG~\cite{ouraccv}. In general, our BiFuse++ achieves state-of-the-art performance under self-supervised scenarios and is comparable with HoHoNet~\cite{hohonet} for supervised training. To investigate the benefit of incorporating cubemap projection, we add rotation noise into both training and testing datasets. BiFuse++ is shown to be robust for panoramic distortion. Furthermore, we estimate the computational cost of different fusion architectures, and our BiFuse++ achieves the lowest inference memory usage among the fusion approaches. In general, BiFuse++ reduces about $80\%$ of parameters compared to BiFuse, while achieving a significantly better performance of depth estimation. Hence, our proposed method is efficient and effective for 360$^\circ$ depth estimation.

To summarize, our contributions are the following:
\begin{enumerate}
    \item BiFuse++ is the first work that integrates the bi-projection fusion architecture into self-supervised monocular 360$^\circ$ depth estimation.
    
    \item We propose a new fusion module to improve the efficiency while preserving complete details in the depth maps.
    
    \item Our framework achieves state-of-the-art performance under self-supervised training scenario and comparable with recent approaches under supervised training.
\end{enumerate}
\section{Related Works}

\noindent \textbf{Supervised Monocular Depth Estimation.~}
Saxena~\etal~\cite{make3d} is the pioneering work lifting a 2D image into a 3D model. With the advance in deep learning, approaches based on convolutional neural networks are studied. Eigen~\etal~\cite{eigen} first propose using a deep neural network to estimate the depth map from a single image. Laina~\etal~\cite{FCRN} adopt ResNet~\cite{ResNet} as the encoder and propose an up-projection module to upsample the feature maps. In addition, reverse Huber loss is proposed to balance the difference between small and large error areas of the estimated depth maps. To further improve the estimated depth maps, several approaches utilizing conditional random field (CRF) are proposed~\cite{Xu_2018_CVPR,depth_classification,Wang_2015_CVPR,Liu_2015_CVPR,Xu_2017_CVPR}. Cao~\etal~\cite{depth_classification} treat depth estimation as a classification task and apply CRF to refine it. By leveraging ordinal regression into the deep neural network, Fu~\etal~\cite{DORN} propose a deep ordinal regression network (DORN) that formulates depth estimation as a classification task. Ranftl~\etal~\cite{midas} propose to combine data from different sources and mix several datasets to greatly improve monocular depth estimation.
Bhat~\etal~\cite{Bhat_2021_CVPR} propose to predict the depth maps as a linear combination of bins. 

However, accurate depth maps measured by laser scanners like LiDARs or Kinect are required for the methods above, which significantly increases the cost of collecting a large amount of training data, and thus self-supervised approaches are studied. \newline

\noindent \textbf{Self-Supervised Monocular Depth Estimation.~}
Xie~\etal~\cite{deep3d} collect the training data from 3D movies, i.e., the left and right frames, and propose Deep3D that infers the left or right frame of the input image to convert a 2D image into a 3D one. Garg~\etal~\cite{garg2016unsupervised} propose a stereopsis-based framework that takes a single image of a rectified stereo pairs as input and infers the corresponding depth maps by image reconstruction error. Godard~\etal~\cite{leftrightconsistency} use the stereo image pairs and left-right photometric consistency to achieve self-training of monocular depth estimation. Since stereo cameras are still less popular than monocular ones for consumer-level devices, self-training approaches using sequential videos are studied. Zhou~\etal~\cite{sfmlearner} propose using two sub-networks to estimate both the monocular depth map and the camera pose of the sequential pairs in the training stage. In this way, a depth estimation network can be directly trained with a large number of monocular videos without any annotation or calibration, which highly improves the scalability of depth estimation. Vijayanarasimhan~\etal~\cite{sfmnet} propose forward-backward constraints and leverage \cite{se3net} to deal with the rigid motions of dynamic objects in the scene. Yang~\etal~\cite{yang2018unsupervised} use depth-normal consistency to improve the depth estimation. Godard~\etal~\cite{godard2019digging} propose an automatic masking technique to efficiently mask out the moving objects. Johnston~\etal~\cite{Johnston_2020_CVPR} use self-attention and discrete disparity to improve depth estimation. Guizilini~\etal~\cite{guizilini20203d} propose PackNet to improve the generalization ability of the depth estimation network on the out-of-domain data. Bian~\etal~\cite{bian2021unsupervised} propose a self-discovered masking scheme to detect moving objects in the videos.

To encode the structural information into the network, object-level information is used to improve the performance of self-supervised depth estimation.
Guizilini~\etal~\cite{Guizilini2020Semantically-Guided} utilize an additional segmentation network to guide the depth estimation network. Chen~\etal~\cite{chen2019towards} propose SceneNet to jointly constrain semantic and geometric understanding with content consistency. Zhu~\etal~\cite{Zhu_2020_CVPR} propose explicit border consistency between segmentation and depth map. Klingner~\etal~\cite{klingner2020eccv} propose SGDepth to solve moving objects by semantic guidance. Hoyer~\etal~\cite{Hoyer_2021_CVPR} transfer the feature maps of a self-supervised depth estimation network to improve semantic segmentation. \newline

\noindent \textbf{360$^\circ$ Perception.~}
Recently, since omnidirectional cameras, e.g., fisheye and 360$^\circ$ cameras, are widely used, people have started to focus on topics of panoramas. To extend the existing deep neural network techniques to panoramas, the distortion introduced by the equirectangular projection increases the instability of performance. Cheng~\etal~\cite{cubepadding} first use cubemap projection to solve the distortion and propose cubepadding to extend the receptive field of each face. Wang~\etal~\cite{wang2018omnidirectional} incorporate circular padding and rotation invariance into the deep neural network. In addition to avoiding the distortion with projections, several distortion-aware convolutional approaches are proposed. Esteves~\etal~\cite{Esteves_2018_ECCV} and Cohen~\etal~\cite{sphericalcnns} propose spherical CNNs by using Fourier transformation to implement the spherical correlation. Su~\etal~\cite{yu-sphericalcnn,KTN} use different convolutional kernels according to the longitude and latitude on the equirectangular projection, and also adapt a pretrained model of perspective camera for inference procedure. \newline

\noindent \textbf{360$^\circ$ Depth Estimation.~}
\revise{Based on the cube padding strategy (Cheng~\etal~\cite{cubepadding}), Wang~\etal~\cite{ouraccv} propose 360-SelfNet that is the first framework of self-supervised 360$^\circ$ depth estimation. Zioulis~\etal~\cite{omnidepth} incorporate SphConv (Su~\etal~\cite{yu-sphericalcnn}) into the encoder to overcome the panoramic distortion and propose a framework for 360$^\circ$ depth estimation. Zioulis~\etal~\cite{sphericalview} use CoordConv (Liu~\etal~\cite{coordconv}) and trinocular view synthesis to improve the performance. Inspired by Yang~\etal~\cite{dula-net} adopting a combination of different projections, Wang~\etal~\cite{bifuse} propose BiFuse that is a two-branch architecture and utilizes cubemap and equirectangular projections. Since the cube padding does not follow the projection geometry, ``spherical padding'' based on the spherical projection is introduced. Moreover, to better leverage the information of two projections, i.e., equirectangular and cubemap, ``bi-projection fusion'' is applied to fuse the feature maps. To improve efficiency, Jiang~\etal~\cite{unifuse} propose UniFuse to simplify the framework of BiFuse. To improve depth prediction, Jin~\etal~\cite{360depth_layout_cvpr} and Zeng~\etal~\cite{360depth_layout_eccv} utilize layout information to provide more context to neural networks. By leveraging 1-D representation proposed in HorizonNet (Sun~\etal~\cite{horizonnet}), Sun~\etal~\cite{hohonet} and Pintore~\etal~\cite{slicenet} propose HoHoNet and SliceNet to train a 360$^\circ$ depth estimation network.}

However, only a few of the abovementioned works try to discuss monocular 360$^\circ$ depth estimation under the self-supervised training scenario. The appropriate and efficient design of networks under such a scenario has not been studied well in the literature. In this paper, we propose ``BiFuse++'', a self-supervised 360$^\circ$ depth estimation framework, to improve the depth estimation efficiency and accuracy. Our framework can be adopted in both supervised and self-supervised training scenarios, and we conduct extensive experiments to verify BiFuse++ under the two scenarios. We will detail our approach in the following sections.

\section{Approach}
\begin{figure*}[t]
    \centering
    \includegraphics[width=\textwidth]{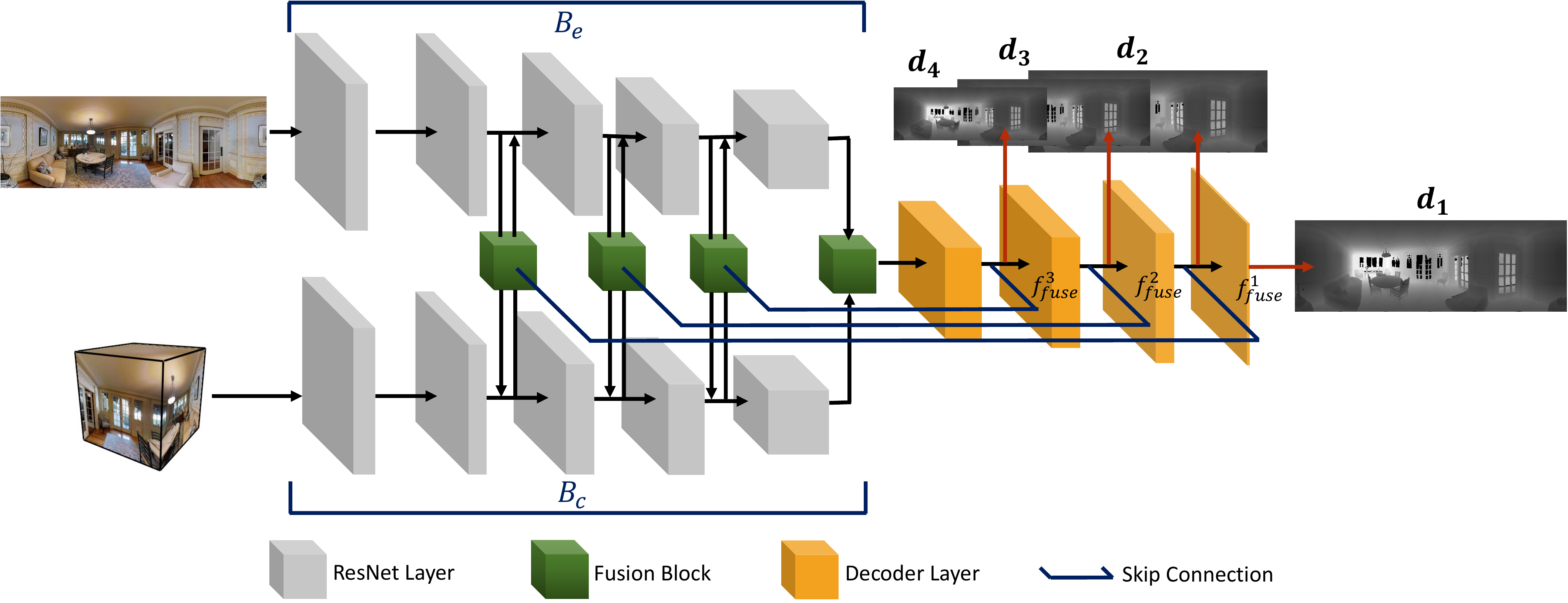}
    \caption{The overview of our DepthNet. Our DepthNet consists of two encoders ($B_e$ and $B_c$) based on \revise{ResNet-34} and a single shared decoder that unifies the feature maps from the two encoders. \revise{The inputs are the equirectangular and cubemap projections converted from a single panorama, and the output is the corresponding equirectangular depth map.} During the encoding procedure, the feature maps of $B_e$ and $B_c$ are fused by our proposed fusion module (green). Unlike \cite{bifuse} and \cite{unifuse}, our fusion module refines the original feature maps and the refined ones are then passed into next layers of $B_e$ and $B_c$. To preserve complete details in the final predicted depth maps, we add three skip-connections by concatenating the fused feature maps ($f_{fuse}^1$, $f_{fuse}^2$, $f_{fuse}^3$) from fusion modules with decoded feature maps. Then, we extract multi-scale depth maps ($d_1$, $d_2$, $d_3$, and $d_4$) from these concatenated feature maps by 1x1 convolutional layers.}
    \label{fig:arch-dispnet}
\end{figure*}
\begin{figure}
    \centering
    \includegraphics[width=\columnwidth]{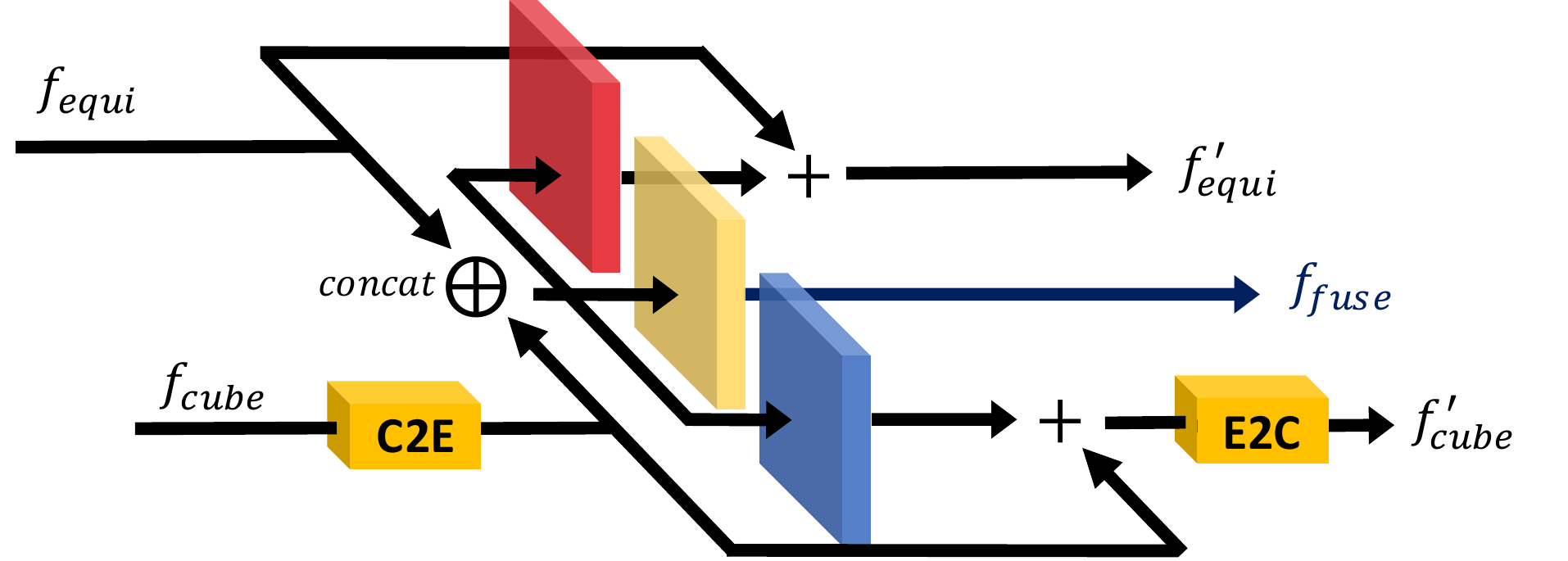}
    \caption{The architecture of our fusion module. The feature maps from equirectangular and cubemap branches are first concatenated and passed into three convolutional blocks. Then, we add a skip connection to the original feature maps and obtain the fused feature maps $f^{\prime}_{equi}$ and $f^{\prime}_{cube}$, which are the inputs of the next convolutional layers. In addition, the other fused feature map $f_{fuse}$ are concatenated in the decoding process later.}
    \label{fig:our-fusion}
\end{figure}

Our BiFuse++ is the first framework that utilizes bi-projection architecture in self-supervised training for monocular 360$^\circ$ depth estimation. In addition, we propose several components to improve the performance and efficiency of BiFuse~\cite{bifuse} and 360-SelfNet~\cite{ouraccv}:

\begin{enumerate}
    \item We propose an advanced bi-projection architecture that significantly reduces the model size while improving depth estimation performance compared with BiFuse.
    \item We propose a new fusion module that is able to effectively share the information between different projections while taking the lowest number of parameters.
    \item To improve the training stability of 360-SelfNet on real-world videos (see Figure~\ref{fig:accv-realworld}), we propose Contrast-Aware Photometric Loss to balance photo consistency error difference between high-texture and low-texture areas.
\end{enumerate}
We first explain the spherical projection and introduce basic transformations in Sec.~\ref{sec:projection}. 
In section~\ref{sec:ours}, we first detail the proposed fusion module of BiFuse++. Then, we introduce the design of our entire framework, including network architectures and the adaptation to supervised and self-supervised training scenarios. Lastly, we explain the proposed Contrast-Aware Photometric Loss and other loss functions we adopted in BiFuse++.

\subsection{Spherical Projection} \label{sec:projection}
\revise{For a cubemap representation of which side length is equal to $w$, we denote $i$ as the six faces $i \in \{B, D, F, L, R, U\}$, to represent the faces of back, down, front, left, right, and up, respectively. Since the field-of-view (FoV) of each face is equal to $90^\circ$; each face can be considered as a perspective camera whose focal length is $\frac{w}{2}$ and all faces share the same center point in world coordinate. Since the viewing direction of each face is fixed in cubemap projection, the corresponding extrinsic matrix of each camera can be defined by a rotation matrix $R_i$. For a pixel $p$ on a certain face $i$, we can transform it into the coordinate on the equirectangular projection by the following mapping:
\begin{equation}
    \begin{aligned}
        &K = \begin{bmatrix}
            w/2 &   0 & w/2 \\
            0 & w/2 & w/2 \\
            0 &   0 &   1
        \end{bmatrix}~, \\
        &q = R_i \cdot K^{-1} \cdot \hat{p}~, \\
        &\theta = \arctan(\frac{q^x}{q^z})~, \\
        &\phi = \arcsin(\frac{q^y}{\abs{q}})~.
    \end{aligned}
\end{equation}
where $w$ is the dimension of $i$ and $\hat{p}$ is the homogeneous representation of $p$; and $\theta$ and $\phi$ are longitude and latitude in equirectangular projection; and $q^x$, $q^y$, $q^z$ are x-y-z components of $q$, respectively. We call such mapping as cube-to-equirectangular (C2E) transformation. Since C2E mapping is reversible, we call the reverse one as equirectangular-to-cube (E2C) transformation. We detail E2C transformation in the following:
\begin{equation}
    \begin{aligned}
        q^x &= \sin(\theta) \cdot \cos(\phi)~, \\
        q^y &= \sin(\phi)~, \\
        q^z &= \cos(\theta) \cdot \cos(\phi)~, \\
        \hat{p} &= K \cdot R_i^T \cdot q~.
    \end{aligned}
\end{equation}
Both E2C and C2E transformation are extensively used in the architecture of BiFuse. For the convenient purpose, we use $\pi(q) = (\theta, \phi)$ and $\pi^{-1}(\theta, \phi) = q$ to represent the forward and inverse spherical projection. To convert an equirectangular depth map to the corresponding point cloud, and project them onto another equirectangular image, we define the following mapping function:
\begin{equation}
    \begin{aligned}
        &\hat{q} = R \cdot d \cdot \pi^{-1}(\theta, \phi) + t~, \\
        &(\hat{\theta}, \hat{\phi}) = \pi(\frac{\hat{q}}{\abs{\hat{q}}})~.
    \end{aligned}
\end{equation}
where $d$ is the depth value, $R$ and $t$ are the camera pose (rotation and translation) of target equirectangular image, and $(\hat{\theta}, \hat{\phi})$ are the projected longitude and latitude on target equirectangular image, respectively.}

\subsection{Our BiFuse++ Framework} \label{sec:ours}

The overview of our BiFuse++ is illustrated in Figure~\ref{fig:teaser}. To achieve self-supervised learning for monocular 360$^\circ$ depth estimation, our training process takes three adjacent panoramas extracted from video sequences, and we adopt two networks, i.e., DepthNet and PoseNet, to estimate the depth map and camera motions. In our DepthNet, we use our proposed fusion module to exchange the information of different projections. With the predicted depth map and camera motions, we propose ``Contrast-Aware Photometric Loss'' to self-supervise the two networks. The details of each component are explained in the following.
\newline

\noindent \textbf{Fusion Module.~} 
The overview of our fusion module is illustrated in Figure~\ref{fig:our-fusion}. Our fusion module consists of three convolutional layers and their inputs are the concatenation of $f_{equi}$ and $f_{cube}$ that are the feature maps of different projections, i.e., equirectangular and cubemap. The red and blue layers are adopted as the residual block to refine the feature maps of different projections, while the last convolution layer (yellow) learns a fused feature map of both projections. Thus, there are three output feature maps, $f_{equi}^\prime$, $f_{cube}^\prime$, and $f_{fuse}$, from our fusion module. Specifically, we generate the three feature maps as:
\begin{equation}
    \begin{aligned}
        f_{equi}^\prime &= f_{equi} + H_e(f_{equi} \oplus C2E(f_{cube}))~, \\
        f_{cube}^\prime &= E2C(C2E(f_{cube}) + H_c(f_{equi} \oplus C2E(f_{cube})))~, \\
        f_{fuse} &= H_f(f_{equi} \oplus C2E(f_{cube}))~,
    \end{aligned}
\end{equation}
where $H_e$, $H_c$, and $H_f$ are convolutional layers, and $\oplus$ is the concatenation operation. $f_{fuse}$ is then leveraged in our decoder. $f_{equi}^\prime$ and $f_{cube}^\prime$ are then passed into the next convolutional layer of the encoder in our network. In this way, the image details can be well preserved in our final predicted depth maps. \newline

\noindent \textbf{Depth Estimation Network (DepthNet).~}
The overview of our DepthNet is illustrated in Figure~\ref{fig:arch-dispnet}. \revise{We take the equirectangular and cubemap projections converted from a single panoramic image as inputs, and our DepthNet predicts the corresponding depth map in equirectangular projection.} DepthNet consists of two encoders based on ResNet-34~\cite{ResNet} to extract the feature maps from equirectangular and cubemap panoramas. We apply our fusion module to fuse the feature maps between each ResNet layer of the two encoders. Different from \cite{unifuse}, we have the refined feature maps from our fusion module ($f_{equi}^\prime$ and $f_{cube}^\prime$ in Figure~\ref{fig:our-fusion}) forwarded into the next layers of our encoders. In this way, the benefits of different projections can be early received in our encoders, and we find that such a mechanism can well preserve the details of panoramas. Similar to \cite{unifuse}, we adopt a single UNet-like decoder to simplify the decoder of BiFuse. We add three skip-connections by concatenating the fused feature maps from our fusion modules ($f_{fuse}$ in Figure~\ref{fig:our-fusion} and $f_{fuse}^{1,2,3}$ in Figure~\ref{fig:arch-dispnet}) with our decoder layers.
In addition, we adopt sub-pixel convolution~\cite{pixelshuffle} as our decoder layers to improve both final accuracy and reduce memory consumption compared to \cite{unifuse} and \cite{bifuse}.
After each decoder layer, we extract the corresponding depth maps of four scales $\{d_s\}_{s=1}^4$ by 1x1 convolutional layers, and we follow \cite{sfmlearner} to add a sigmoid layer after them, with $\alpha$ and $\beta$ to control the range of the depth value.
\begin{equation}
    \begin{aligned}
        d_s^\prime = \alpha & \cdot \rho (f_s) + \beta~, \\
        d_s &= \frac{1}{d_s^\prime}~.
        \label{eq:alpha-lifting}
    \end{aligned}
\end{equation}
where $s$ denotes the scale, $f_s$ is the output of the four convolutional layers, $\alpha$ and $\beta$ are hyper-parameters, $\rho$ is the sigmoid function, and $d_s$ is the depth map of scale $s$. In this paper, we follow \cite{sfmlearner} and set $\alpha=10$ and $\beta=0.01$. \newline

\begin{figure}
    \centering
    \includegraphics[width=\columnwidth]{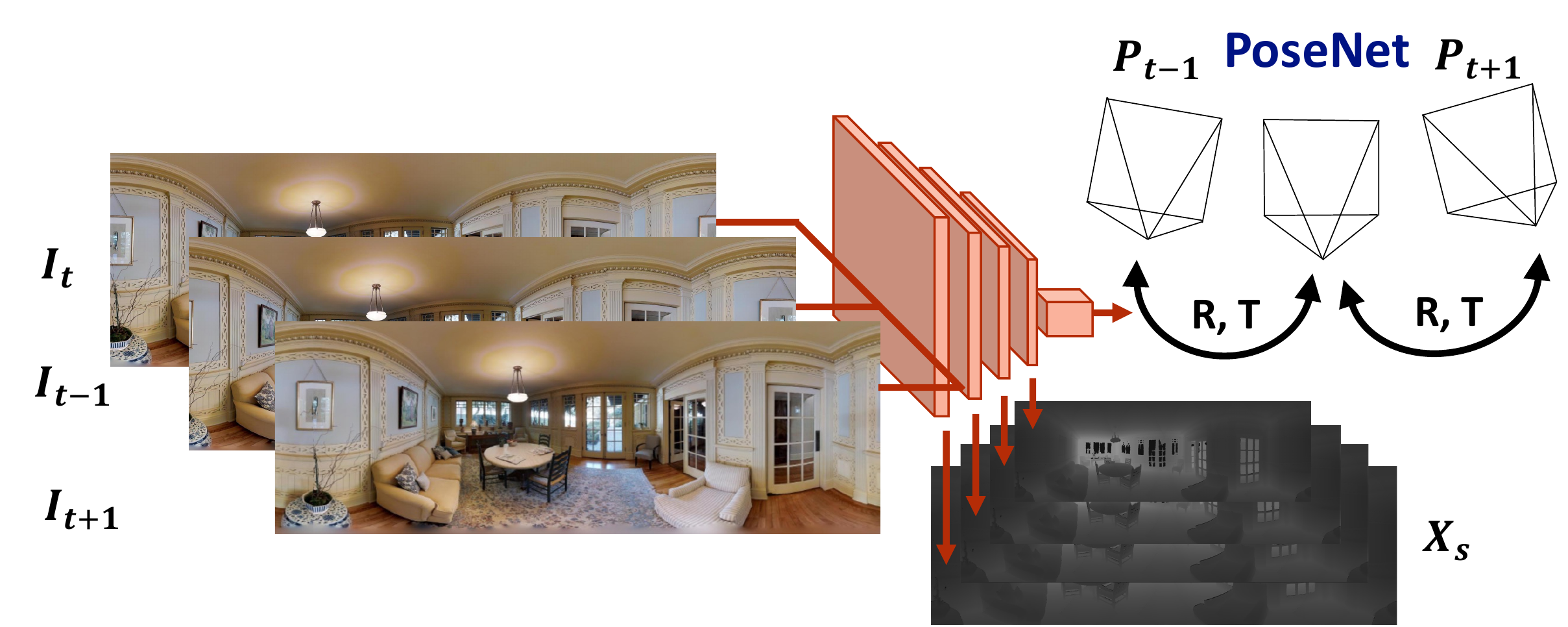}
    \caption{\revise{The overview of our PoseNet. Our PoseNet is based on ResNet-18 and the inputs are three sequential panoramas ($I_{t-1}, I_t, I_{t+1}$) in a video and PoseNet infers the corresponding camera motion $P_{t-1}$ and $P_{t+1}$. To suppress the ambiguity of photo consistency error in occluded areas and stabilize the training, PoseNet estimates four occlusion masks $X_s$ to find the occluded areas.}}
    \label{fig:posenet}
\end{figure}

\noindent \textbf{Pose Estimation Network (PoseNet).~}
To achieve self-supervised depth estimation on videos, both depth and camera motion are required to estimate photo consistency errors. Hence, we adopt an addition network (PoseNet) to estimate the corresponding camera motions between adjacent frames. As illustrated in Figure~\ref{fig:posenet}, we adopt a single ResNet-18~\cite{ResNet} encoder of which inputs are the concatenation of three sequential panoramas ($I_{t-1}$, $I_t$, and $I_{t+1}$). We estimate the backward and forward camera motions $P_{t-1}$ and $P_{t+1}$ to jointly calculate their photo consistency errors. Since photo consistency has ambiguity in occluded areas; we have four additional 3x3 convolutional layers that take the four feature maps from ResNet-18 as inputs and estimate the occlusion masks at four scales, which are denoted as $\{X_s\}_{s=1}^4$. We then use the occlusion masks to suppress ambiguous areas and stabilize the training. \newline

\noindent \textbf{Self-supervised Loss Function.~}
We use the photo consistency assumption, i.e., the image intensity is consistent across reprojected frames given the depth and camera motion, to train our DepthNet and PoseNet in a self-supervised fashion. Regarding the loss function, 360-SelfNet~\cite{ouraccv} first proposes ``Spherical Photometric Loss (SPL)'' to calculate the photo consistency error with spherical projection. However, 360-SelfNet cannot estimate stable depth maps on low-texture areas in real-world videos (Figure~\ref{fig:accv-realworld}). We find that there is a degeneration problem in SPL, i.e., the lower SPL is not always equal to more accurate depth maps, and we show supporting evidence in Section~\ref{sec:experiments}. The degeneration comes from the ambiguity of photo consistency assumption that the consistency errors of pixels in low-texture areas are meaningless; such ambiguity can seriously harm the training results in real-world videos. To this end, we propose ``Contrast-Aware Photometric Loss (CAPL)'' to prevent networks from being affected by these low-texture areas:
\begin{equation}
    \begin{aligned}
        &\mathcal{L}_{CAPL}^s(I_{t,s}) = \sum_{p=1}^N X_{s}(p) \cdot \sigma(I_{t,s}(p)) \cdot \delta(I_{t,s}(p)) ~,\\
        &\delta(I_{t,s}(p)) = \abs{I_{t,s}(p) - \hat{I}_{t-1,s}(p)} + \abs{I_{t,s}(p) - \hat{I}_{t+1,s}(p)}~,
    \end{aligned}
\end{equation}
where $p$ denotes pixels, $N$ is the total number of pixels, $X_s$ is the predicted occlusion mask, and $\sigma$ is the standard deviation of $p$ in a 5x5 window. $\delta$ is the photo consistency error of $I_{t,s}$; $\hat{I}_{t-1,s}$ and $\hat{I}_{t+1,s}$ are the warped panoramas of $I_{t,s}$ after being reprojected onto $I_{t-1,s}$ and $I_{t+1,s}$ by the predicted depth map $d_s$ and camera motions ($P_{t-1}$ and $P_{t+1}$), respectively. $\delta$ is first multiplied by the occlusion mask to remove unreasonable depth values, and then we use the standard deviation of $p$ to solve the degeneration problem since the standard deviation in low-texture areas are usually small. In this way, CAPL can significantly improve the quality of predicted depth maps in real-world videos.

In addition to CAPL, we adopt two regularization terms to provide constraints for occlusion masks and predicted depth maps. To prevent predicted occlusion masks from decaying to zero, we apply a binary cross-entropy loss to the masks:
\begin{equation}
    \begin{aligned}
        \mathcal{L}_m (X_s) = -\sum_{p=1}^N \log(X_s(p))~,
    \end{aligned}
\end{equation}
This regularization provides a large penalty when the values in occlusion masks are small. To reduce the noise on the predicted depth maps, we apply smooth regularization to the predicted depth map:
\begin{equation}
    \begin{aligned}
        \mathcal{L}_{sm}(d_s) = \sum_{p=1}^N \abs{\nabla(d_s(p))}~.
    \end{aligned}
\end{equation}

\noindent \textbf{Supervised and Self-Supervised Training.~}
In this paper, our proposed framework is evaluated under both supervised and self-supervised training scenarios. In supervised training, our DepthNet is directly trained with ground truth depth maps, and we adopt reverse Huber loss~\cite{FCRN} as our loss function $\mathcal{L}_{berHu}$:
\begin{align}
    \label{eq:berhu}
    &\mathcal{L}_{berHu} = \sum_{s=1}^4 \sum_{p=1}^N \mathcal{B}(d_s(p), \hat{d}_s(p)) ~,\\
    \mathcal{B}(d_s(p), \hat{d}_s(p)) &= 
    \left\{
    \begin{array}{lr}
         |d_s(p) - \hat{d}_s(p)| & |d_s(p) - \hat{d}_s(p)| \leq c ~,\\
         \frac{(d_s(p) - \hat{d}_s(p))^2 + c^2}{2c} & |d_s(p) - \hat{d}_s(p)| > c~,
    \end{array}
    \right.
\end{align}
where $p$ is pixels, while $d_s$ and $\hat{d}_s$ are the predicted and ground truth depth maps, respectively. $c$ is typically set to $0.2 \cdot \max(|d_s(p) - \hat{d}_s(p)|)$.

In self-supervised training, DepthNet and PoseNet are trained with three abovementioned loss terms: 1) Contrast-Aware Photometric Loss, 2) occlusion mask regularization, and 3) smooth regularization. The final loss function is then established as:
\begin{equation}
    \begin{aligned}
        \mathcal{L}_{ss} = \sum_{s=1}^4 \mathcal{L}_{CAPL}^s(I_{t,s}) + w_1 \cdot \mathcal{L}_m (X_s) + w_2 \cdot \mathcal{L}_{sm}(d_s)~,
    \end{aligned}
    \label{eq:self-loss}
\end{equation}
where $w_1$ and $w_2$ are hyper-parameters.
\section{Experimental Results} \label{sec:experiments}

\begin{table*}[htbp]
  \centering
  \caption{The quantitative results on Matterport3D~\cite{mp3d}.}
  \resizebox{0.8\textwidth}{!}{\begin{tabular}{|c|c|c|c|c|c|c|c|}
    \hline
    \textbf{Method} & \textbf{MAE} $\downarrow$ & \textbf{MRE} $\downarrow$ & \textbf{RMSE} $\downarrow$ & \textbf{RMSE (log)} $\downarrow$ & {\boldmath$\delta_1$ $\uparrow$} & {\boldmath$\delta_2$ $\uparrow$} & {\boldmath$\delta_3$ $\uparrow$}  \bigstrut\\
    \hline
    \hline
    FCRN  & 0.4008 & 0.2409 & 0.6704 & 0.1244 & 0.7703 & 0.9174 & 0.9617 \\
    OmniDepth & 0.4838 & 0.2901 & 0.7643 & 0.1450 & 0.6830 & 0.8794 & 0.9429 \\
    BiFuse & 0.3470 & 0.2048 & 0.6259 & 0.1134 & 0.8452 & 0.9319 & 0.9632 \\
    UniFuse & 0.3160 & 0.1592 & 0.5485 & 0.0926 & 0.8490 & 0.9463 & 0.9747 \\
    SliceNet & 0.3296 & 0.1764 & 0.6133 & 0.1045 & 0.8716 & 0.9483 & 0.9716 \\
    HoHoNet & 0.2862 & 0.1488 & \textbf{0.5138} & 0.0871 & 0.8786 & \textbf{0.9519} & 0.9771 \\
    \textbf{BiFuse++} & \textbf{0.2842} & \textbf{0.1424} & 0.5190 & \textbf{0.0862} & \textbf{0.8790} & 0.9517 & \textbf{0.9772} \bigstrut[b]\\
    \hline
    \end{tabular}}%
    \label{tab:mp3d}%
\end{table*}%
\begin{table*}[htbp]
  \centering
  \caption{The quantitative results on Stanford2D3D~\cite{stfd}. Note that SliceNet$^*$ is a re-implemented version.}
  \resizebox{0.8\textwidth}{!}{\begin{tabular}{|c|c|c|c|c|c|c|c|}
    \hline
    \textbf{Method} & \textbf{MAE} $\downarrow$ & \textbf{MRE} $\downarrow$ & \textbf{RMSE} $\downarrow$ & \textbf{RMSE (log)} $\downarrow$ & {\boldmath$\delta_1$ $\uparrow$} & {\boldmath$\delta_2$ $\uparrow$} & {\boldmath$\delta_3$ $\uparrow$}  \bigstrut\\
    \hline
    \hline
    FCRN  & 0.3428 & 0.1837 & 0.5774 & 0.1100 & 0.7230 & 0.9207 & 0.9731 \\
    OmniDepth & 0.3743 & 0.1996 & 0.6152 & 0.1212 & 0.6877 & 0.8891 & 0.9578 \\
    BiFuse & 0.2343 & 0.1209 & 0.4142 & 0.0787 & 0.8660 & 0.9580 & 0.9860 \\
    UniFuse & 0.2198 & 0.1195 & 0.3875 & 0.0747 & 0.8686 & 0.9621 & 0.9870 \\
    SliceNet$^*$ & 0.2484 & 0.1249 & 0.4370 & 0.0873 & 0.8377 & 0.9414 & 0.9777 \\
    HoHoNet & \textbf{0.2027} & \textbf{0.1014} & 0.3834 & \textbf{0.0668} & \textbf{0.9054} & \textbf{0.9693} & \textbf{0.9886} \\
    \textbf{BiFuse++} & 0.2173 & 0.1117 & \textbf{0.3720} & 0.0727 & 0.8783 & 0.9649 & 0.9884 \bigstrut[b]\\
    \hline
    \end{tabular}}%
    \label{tab:stanford2d3d}%
\end{table*}%
  
\begin{table*}[thbp]
    \centering
    \caption{The quantitative results on PanoSUNCG~\cite{ouraccv}.}
    \resizebox{0.8\textwidth}{!}{\begin{tabular}{|c|c|c|c|c|c|c|c|}
      \hline
      \textbf{Method} & \textbf{MAE} $\downarrow$ & \textbf{MRE} $\downarrow$ & \textbf{RMSE} $\downarrow$ & \textbf{RMSE (log)} $\downarrow$ & {\boldmath$\delta_1$ $\uparrow$} & {\boldmath$\delta_2$ $\uparrow$} & {\boldmath$\delta_3$ $\uparrow$}  \bigstrut\\
      \hline
      \hline
      FCRN  & 0.1346 & 0.0979 & 0.3973 & 0.0692 & 0.9223 & 0.9659 & 0.9819 \bigstrut[t]\\
      OmniDepth & 0.1624 & 0.1143 & 0.3710 & 0.0882 & 0.8705 & 0.9365 & 0.9650 \\
      BiFuse & 0.0789 & 0.0592 & 0.2596 & 0.0443 & 0.9590 & 0.9823 & 0.9907 \\
      UniFuse & 0.0776 & 0.0528 & 0.2704 & 0.0441 & 0.9591 & 0.9825 & 0.9906 \\
      \textbf{BiFuse++} & \textbf{0.0688} & \textbf{0.0524} & \textbf{0.2477} & \textbf{0.0414} & \textbf{0.9630} & \textbf{0.9835} & \textbf{0.9911} \bigstrut[b]\\
      \hline
      \end{tabular}}%
    \label{tab:panosuncg}%
  \end{table*}%

We first introduce the common evaluation metrics, the benchmark datasets (Sec.~\ref{sec:metric-and-datasets}), and implementation details (Sec.~\ref{sec:imp}). For performance evaluation, we validate the improved accuracy of BiFuse++ network architecture with supervised training scenarios (Sec.~\ref{sec:results-supervised}) on three datasets: Matterport3D~\cite{mp3d}, Stanford2D3D~\cite{stfd}, and PanoSUNCG~\cite{ouraccv}. 
We further test the robustness of our method by evaluating the performance under rotation noise.
Moreover, we validate the computational efficiency of BiFuse++ with respect to existing approaches (Sec.~\ref{sec:computation}).
Then, we use the BiFuse++ network with low inference memory to conduct self-training efficiently on PanoSUNCG~\cite{ouraccv} (Sec.~\ref{sec:results-selfsupervised}). Moreover, we also capture several videos in the real-world environment and conduct qualitative comparisons to show the applicability of BiFuse++.


\subsection{Evaluation Metrics and Datasets}
\label{sec:metric-and-datasets}
 
We use standard evaluation protocols in depth estimation, i.e., MAE (mean absolute error), MRE (mean relative error), RMSE (root mean square error), RMSE (log) (scale-invariant root mean square error), and $\delta$ (threshold). For RMSE (log), we use log-10 for the computation. During evaluations, we follow \cite{bifuse} to ignore the area in which ground truth depth values are larger than 10 meters. Since the scale of self-training results is ambiguous, we follow \cite{sfmlearner} and apply median alignment before evaluation for self-supervised scheme:
\begin{equation}
    \begin{aligned}
        d^\prime = d \cdot \frac{median(\hat{d})}{median(d)}~.
    \end{aligned}
\end{equation}
where $d$ is the predicted depth map, $\hat{d}$ is the ground truth one, and $d^\prime$ is the median-aligned depth map used for evaluation. \newline

The following datasets are used in our experiments. \newline
\noindent \textbf{Matterport3D.~}
Matterport3D contains 10,800 panoramas and the corresponding depth ground truth captured by Matterport’s Pro 3D Camera, a structured-light scanner. This dataset is the largest real-world dataset for indoor panorama scenes. However, the depth maps from sensors usually have noise or missing value in certain areas. In practice, we filter areas with missing values during training. We follow the official split to train and test our network, which takes 61 rooms for training and the others for testing. We resize the resolution of images and depth maps into 512 $\times$ 1024. \newline

\noindent \textbf{Stanford2D3D.~}
Stanford2D3D is collected from three kinds of buildings in the real world, containing six large-scale indoor areas. The dataset contains 1413 panoramas, and we use one of the official splits that takes the fifth area (area 5) for testing, and the others are for training. During training and testing, we resize the resolution of images and depth maps into 512 $\times$ 1024. \newline

\noindent \textbf{PanoSUNCG.~}
PanoSUNCG contains 103 scenes of SunCG~\cite{suncg} and has 25,000 panoramas. In our experiments, we use the official training and testing splits, where 80 scenes are for training and 23 for testing. For the supervised scheme, we resize them to 256 $\times$ 512 and filter out pixels with depth values larger than 10 meters. For the self-supervised one, we keep the original resolution 512 $\times$ 1024.

\subsection{Implementation Details}
\label{sec:imp}
We implement our network using PyTorch~\cite{pytorch}. We use Adam~\cite{adam} optimizer with $\beta_1=0.9$ and $\beta_2=0.999$. Our batch size is 8, and the learning rate is 0.0003. Unlike BiFuse~\cite{bifuse}, training the network and fusion module separately, we train the entire framework jointly. The ResNet encoders of DepthNet and PoseNet are first pretrained on ImageNet~\cite{imagenet}, and we apply uniform initialization to all other layers. We train the networks for 100 epochs for supervised scenarios, while the networks are trained for 60 epochs for self-supervised scenarios. Following \cite{ouraccv}, we set $w_1$ and $w_2$ of Equation~\eqref{eq:self-loss} to $0.1$ and $0.01$, respectively. \newline

\noindent \textbf{Supervised and Self-supervised Training.~}
\revise{In the supervised training scenario, we directly use the monocular images and corresponding depth maps provided by the above-mentioned datasets to train our DepthNet. In other words, the training is similar to monocular depth estimation like BiFuse~\cite{bifuse}. Since we can directly acquire supervision from ground truth depth maps for training DepthNet with Equation~\eqref{eq:berhu}, our PoseNet is not involved in this scenario. For the self-supervised scenario, since we cannot access ground truth depth maps for training, we use the RGB video sequences from PanoSUNCG~\cite{ouraccv} to self-supervise both DepthNet and PoseNet. Specifically, PoseNet takes three sequential panoramas ($I_{t-1}$, $I_t$, $I_{t+1}$ in Figure~\ref{fig:posenet}) as input and infers the camera motions between them, while DepthNet takes only $I_t$ as input and predicts the corresponding depth map. With the predicted monocular depth map and camera motions, we can self-supervise the two networks with Equation~\eqref{eq:self-loss}.}

\begin{figure*}
    \centering
    \includegraphics[width=\textwidth]{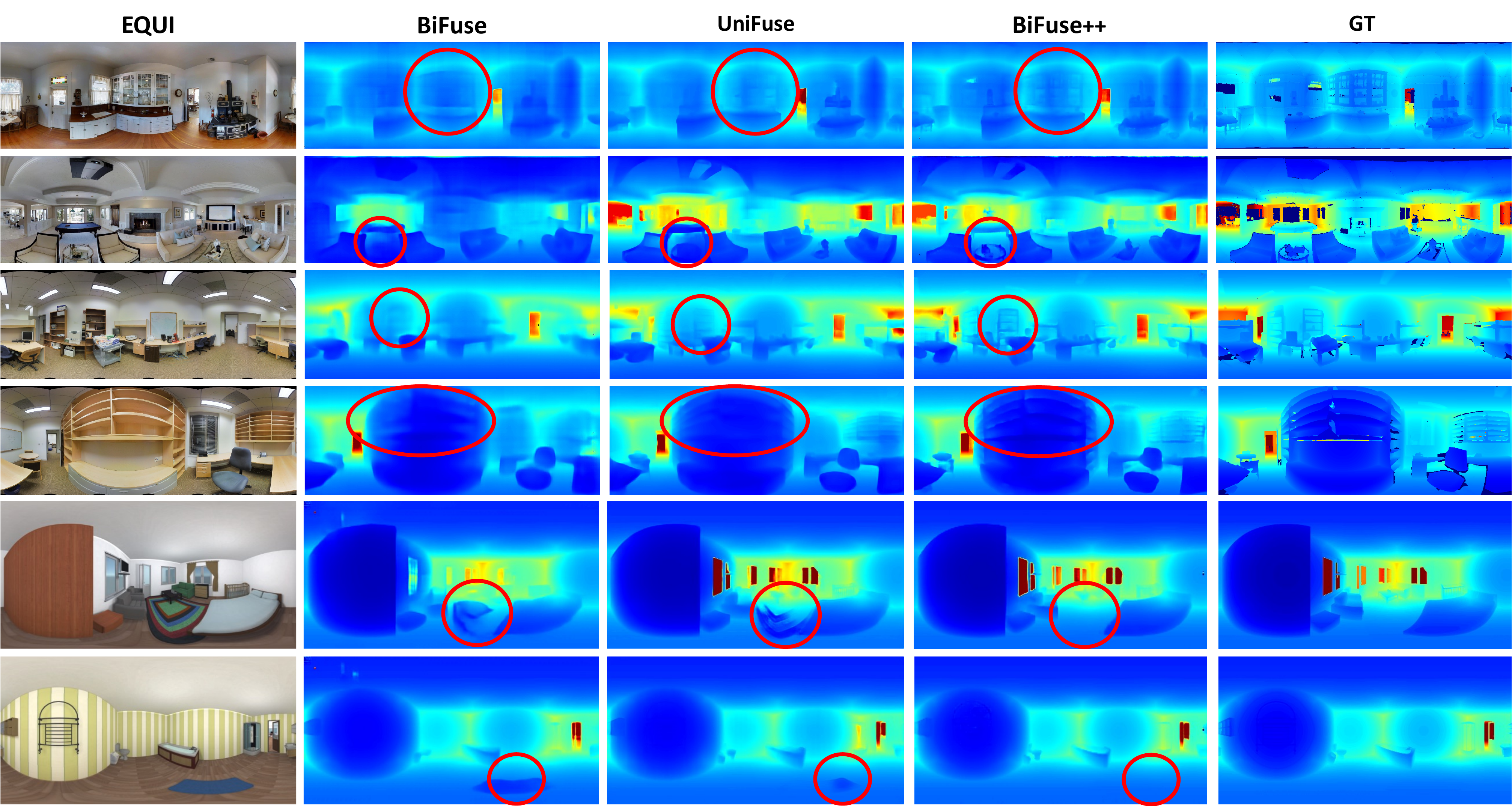}
    \caption{The qualitative results on Matterport3D, Stanford2D3D, and PanoSUNCG under the supervised scenario (every two rows show qualitative results of each dataset). Note that the dark blue and red colors indicate close and far distance, and we use red circles to highlight the inconsistent predictions of all approaches.}
    \label{fig:qualitative-all}
\end{figure*}
\subsection{Results of Supervised Scenario}
\label{sec:results-supervised}
For supervised scenarios, we compare BiFuse++ with works of monocular depth estimation, including approaches designed for both perspective and spherical cameras. 1) FCRN~\cite{FCRN}, a strong approach designed for perspective cameras. 2) OmniDepth~\cite{omnidepth}, a framework designed for spherical cameras that incorporates \cite{yu-sphericalcnn} into the architecture.
For 1-D representation approaches, we compare our method with SliceNet~\cite{slicenet} and HoHoNet~\cite{hohonet}.

The quantitative results are shown in Table~\ref{tab:mp3d}-\ref{tab:panosuncg}. BiFuse++ achieves comparable results with the latest state-of-the-art 1D-representation HoHoNet and outperforms all other methods under the assumption that all input images are well aligned with the gravity direction.

\begin{figure}[t]
    \centering
    \includegraphics[width=\columnwidth]{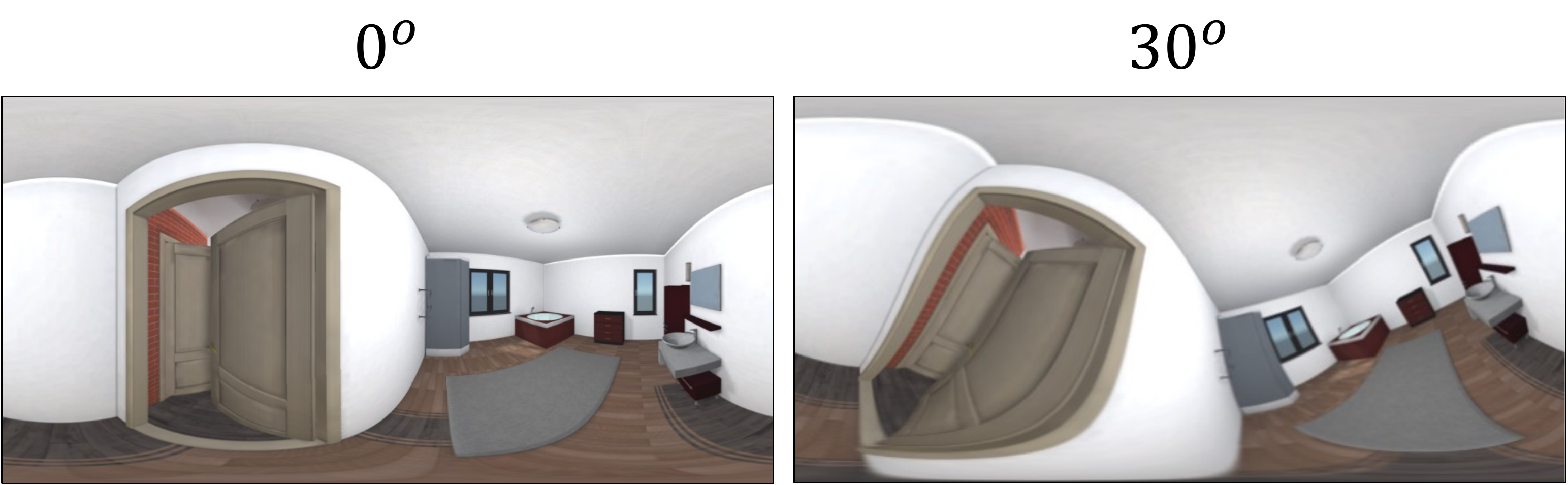}
    \caption{The distortion introduced by equirectangular projection. When the pitch of the camera is 0$^\circ$, the structure of the room is clear. As the pitch becomes larger, the effect of equirectangular distortion is more obvious. The distortion affects the training stability when applying existing approaches designed for perspective cameras to panoramas.}
    \label{fig:distortion}
\end{figure}

\noindent \textbf{Rotation Noise Evaluation.} 
Since the spherical cameras adopted by the abovementioned datasets are well aligned with the gravity direction, the rotation of panoramas is usually small.
This assumption benefits the most for 1-D representation like HoHoNet. To investigate the result without this assumption, we conduct the following experiment. For training and testing, we introduce rotation noise (see $0^\circ$ and $30^\circ$ rotation in Figure~\ref{fig:distortion}) on the Matterport3D and Stanford2D3D datasets. Specifically, we uniformly sample an angle between $30^\circ$ and $-30^\circ$ to rotate panoramas and ground truth depth maps during each training iteration. For testing sets, we also apply the rotation noise, but the rotated angles of each image are consistent across all baselines. BiFuse++ is robust to rotation noise compared to other methods, as shown in Table~\ref{tab:mp3d-rotate}-\ref{tab:stanford2d3d-rotate}.

\begin{figure*}
    \centering
    \includegraphics[width=\textwidth]{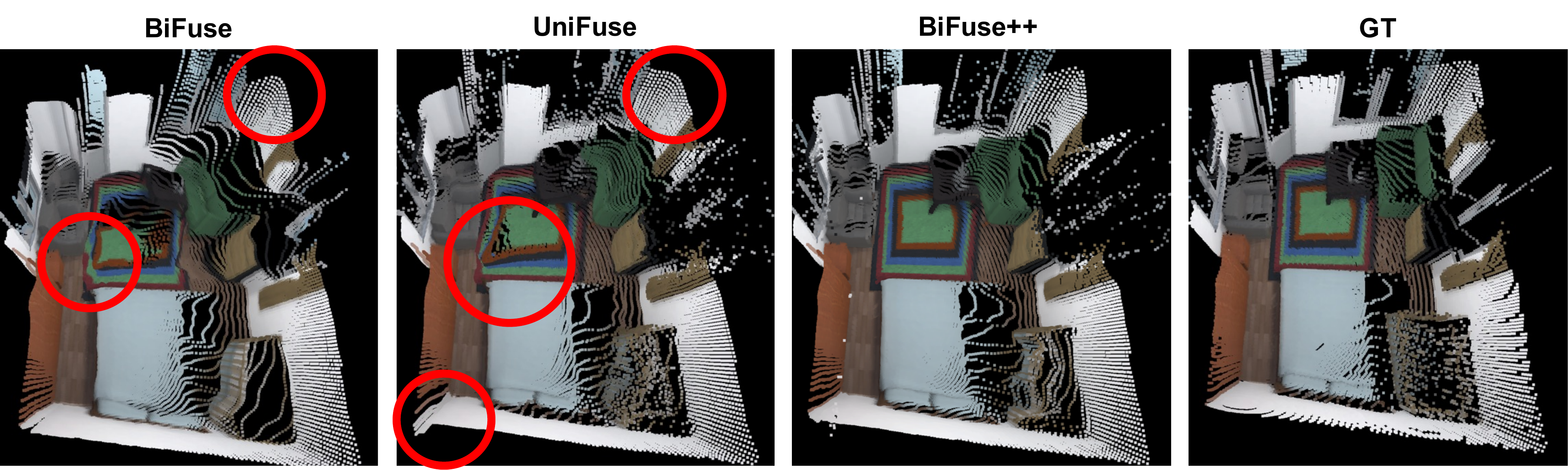}
    \caption{The 3D reconstruction comparison of BiFuse++ with other baselines. We note that the red circles indicate the incorrect depth prediction. Our BiFuse++ is able to preserve the corner details, while the other approaches predict inconsistent results.}
    \label{fig:qualitative-3d}
\end{figure*}

\begin{table}[ht]
  \centering
  \caption{The quantitative results after applying rotation noise on Matterport3D~\cite{mp3d}.}
  \begin{tabular}{|c|c|c|c|c|}
    \hline
    \textbf{Method} & \textbf{MAE} & \textbf{MRE} & \textbf{RMSE} & \textbf{RMSE (log)} \bigstrut\\
    \hline
    \hline
    UniFuse & 0.3197 & 0.1627 & 0.5456 & 0.0941 \bigstrut[t]\\
    SliceNet & 0.3669 & 0.1863 & 0.6124 & 0.1058 \\
    HoHoNet & 0.3085 & \textbf{0.1486} & 0.5385 & \textbf{0.0897} \\
    \textbf{BiFuse++} & \textbf{0.3054} & 0.1521 & \textbf{0.5293} & \textbf{0.0897} \bigstrut[b]\\
    \hline
    \end{tabular}%
    \label{tab:mp3d-rotate}%
\end{table}%
\begin{table}[ht]
  \centering
  \caption{The quantitative results after applying rotation noise on Stanford2D3D~\cite{stfd}.}
  \begin{tabular}{|c|c|c|c|c|}
    \hline
    \textbf{Method} & \textbf{MAE} & \textbf{MRE} & \textbf{RMSE} & \textbf{RMSE (log)} \bigstrut\\
    \hline
    \hline
    UniFuse & 0.2542 & 0.1289 & 0.4209 & 0.0819 \bigstrut[t]\\
    SliceNet & 0.2892 & 0.1459 & 0.5038 & 0.1001 \\
    HoHoNet & 0.2210 & \textbf{0.1116} & 0.4001 & 0.0744 \\
    \textbf{BiFuse++} & \textbf{0.2193} & 0.1134 & \textbf{0.3890} & \textbf{0.0742} \bigstrut[b]\\
    \hline
    \end{tabular}%
    \label{tab:stanford2d3d-rotate}%
\end{table}%

\noindent \textbf{Qualitative Discussion.} 
The qualitative comparison of fusion approaches are shown in Figure~\ref{fig:qualitative-all}. Compared with BiFuse~\cite{bifuse}, BiFuse++ achieves sharper results. This is because BiFuse adopts a simple architecture without any skip-connection layer inherited from FCRN~\cite{FCRN} so that the image details encoded in the low-level feature maps cannot be well preserved. In contrast, Bifuse++ adopts a UNet-like architecture with three skip-connection layers (c.f., Figure~\ref{fig:arch-dispnet}), and the details are well recovered in the depth maps. Compared with UniFuse~\cite{unifuse}, BiFuse++ recovered much clearer object boundaries. This is because UniFuse adopts two ResNet to encode features maps of equirectangular and cubemap projections independently so that the two encoders cannot effectively leverage the information from the other branch. In contrast, we pass the fused features ($f^\prime_{equi}$ and $f^\prime_{cube}$ in Figure~\ref{fig:our-fusion}) to the layers of the two encoders. In this way, the layers of encoders can directly retrieve the context from the other projection and preserve more details on the predicted depth maps eventually. \newline



\noindent \textbf{3D Comparison.~}
To further show the difference in depth maps generated from fusion approaches, we show the corresponding point cloud visualizations in Figure~\ref{fig:qualitative-3d}. The point clouds of BiFuse and UniFuse are not capable of generating sharp corners, while BiFuse++ predicts accurate wall boundaries. Moreover, the depth of objects like carpets is noisier than the one of BiFuse++. For high variance areas like the edges between windows and walls, the results of BiFuse++ are closer to the ground truth. In contrast, the other baselines generate smooth results in these areas. Hence, our BiFuse++ is able to predict accurate point clouds and outperform the other baselines. \newline

\subsection{Computational Comparison}
\label{sec:computation}

Before we apply BiFuse++ to self-training of depth estimation, we first examine the efficiency of different fusion approaches since the self-training procedure usually takes more resources and, thus, a memory-efficient framework is necessary. We estimate the efficiency with the following two aspects.

\noindent\textbf{Number of Parameters.} 
Since the number of parameters used by fusion modules (\cite{bifuse}, \cite{unifuse}, and BiFuse++) depends on the channels of input features, we fix the channel numbers of all approaches to 512 to fairly compare the module size of different fusion modules. As shown in Table~\ref{tab:fusion-params}, the fusion module of BiFuse takes the largest number of parameters, and BiFuse++ uses the smallest one. Under this setting, our new fusion module can reduce $55\%$ of parameters than BiFuse does.

\begin{table}[tbh]
    \centering
    \caption{The number of parameters of different fusion modules (we set the channels to 512).}
      \begin{tabular}{|c|c|c|c|}
      \hline
            & \textbf{BiFuse++} & \textbf{UniFuse} & \textbf{BiFuse} \bigstrut\\
      \hline
      \hline
      \textbf{Parameters} & 2.1 M & 3.5 M & 4.7 M \bigstrut\\
      \hline
      \end{tabular}%
    \label{tab:fusion-params}%
  \end{table}%
  
\begin{table}[tbh]
    \centering
    \caption{The computational comparison of fusion approaches.}
      \begin{tabular}{|c|c|c|c|}
      \hline
      \textbf{Approach} & \textbf{GFLOPs} & \textbf{Model size} & \textbf{Runtime Memory} \bigstrut\\
      \hline
      \hline
      BiFuse++  & 87.42 & 53.19 M & 1762 Mb \bigstrut\\
      \hline
      UniFuse & 62.58 & 30.26 M & 2006 Mb \bigstrut\\
      \hline
      BiFuse & 682.86 & 253.1 M & 3346 Mb \bigstrut\\
      \hline
      \end{tabular}%
      \label{tab:gflops}%
  \end{table}%

\begin{table*}[th]
  \centering
  \caption{The quantitative results on PanoSUNCG~\cite{ouraccv} under the self-supervised scenario. Our BiFuse++ with Spherical Photometric Loss (SPL)~\cite{ouraccv} or Contrast-Aware Photometric Loss (CAPL) outperforms other baselines. SSIM stands for structural similarity.}
    \resizebox{0.8\textwidth}{!}{\begin{tabular}{|c|c|c|c|c|c|c|c|}
    \hline
    \textbf{Method} & \textbf{MAE} $\downarrow$ & \textbf{MRE} $\downarrow$ & \textbf{RMSE} $\downarrow$ & \textbf{RMSE (log)} $\downarrow$ & {\boldmath$\delta_1$ $\uparrow$} & {\boldmath$\delta_2$ $\uparrow$} & {\boldmath$\delta_3$ $\uparrow$} \bigstrut\\
    \hline
    \hline
     360-SelfNet (Equi) & 0.2436 & 0.1499 & 0.5421 & 0.0959 & 0.8618 & 0.9463 & 0.9714 \bigstrut[t]\\
     360-SelfNet & 0.2344 & 0.1521 & 0.5121 & 0.0934 & 0.8479 & 0.9420 & 0.9726 \\
    UniFuse & 0.2452 & 0.1458 & 0.4978 & 0.0920 & 0.8513 & 0.9398 & 0.9691 \\
    BiFuse++ w/ SSIM & 0.3125 & 0.1852 & 0.5889 & 0.1068 & 0.7847 & 0.9313 & 0.9684 \\
    BiFuse++ w/ SPL & 0.2083 & 0.1287 & 0.4695 & 0.0838 & 0.8838 & \textbf{0.9583} & \textbf{0.9778} \\
    BiFuse++ w/ CAPL & \textbf{0.1815} & \textbf{0.1176} & \textbf{0.4321} & \textbf{0.0790} & \textbf{0.8974} & 0.9546 & 0.9773 \bigstrut[b]\\
    \hline
    \end{tabular}}%
  \label{tab:self-panosuncg}%
\end{table*}%

\noindent\textbf{Runtime Resources.} To examine the computational resources of different fusion approaches, we adopt RTX2080Ti as the platform and use a single dummy image of resolution 512x1024x3 as the input of all tested frameworks. The results are shown in Table~\ref{tab:gflops}. Compared to BiFuse, we reduce $87\%$ of GFLOPs and $79\%$ of parameters. Moreover, BiFuse++ only needs half of the inference memory as BiFuse. 
Although our GFLOPs and parameters are slightly larger than UniFuse, we use less inference memory since we adopt PixelShuffle~\cite{pixelshuffle} in the decoding process. In general, we significantly reduce the computational resources of BiFuse and make it more practical to be applied in self-training. 

\begin{figure*}[th]
    \centering
    \includegraphics[width=\textwidth]{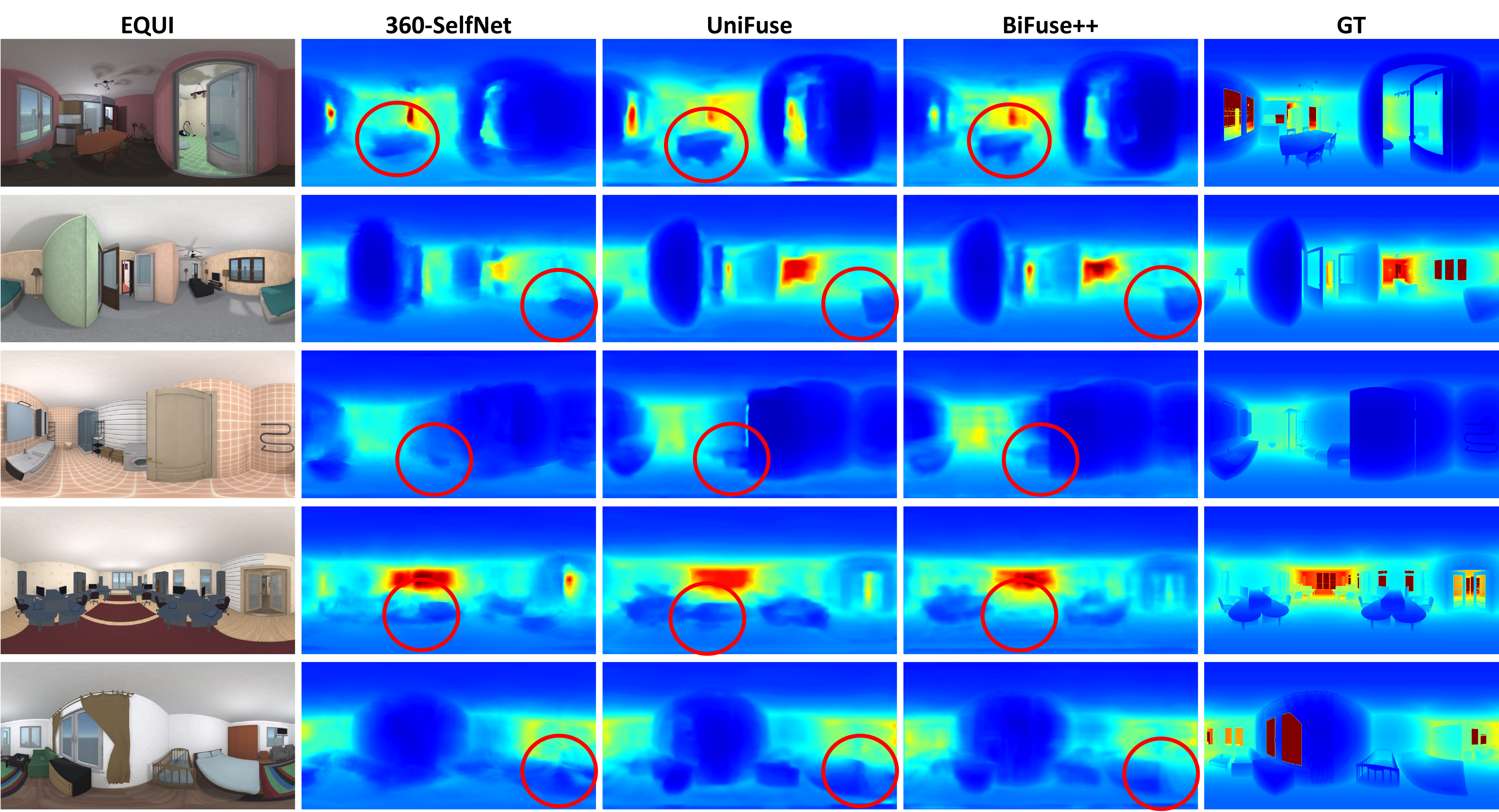}
    \caption{The qualitative results on PanoSUNCG under self-supervised scenario.}
    \label{fig:qualitative-self-panosuncg}
\end{figure*}

\subsection{Results of Self-Supervised Scenario}
\label{sec:results-selfsupervised}

We use the framework described in Section~\ref{sec:ours} to self-supervisedly train our DepthNet and PoseNet. For the qualitative and quantitative comparison, we conduct experiments on PanoSUNCG~\cite{ouraccv} to verify the applicability of BiFuse++. We compare BiFuse++ with three baselines: 1) 360-SelfNet (Equi): the framework of \cite{ouraccv}, but the inputs are equirectangular panoramas. 2) 360-SelfNet: the framework proposed in \cite{ouraccv}. 3) UniFuse: replace our DepthNet with the architecture of \cite{unifuse}. In addition, we compare variants of BiFuse++ with three different loss functions: 1) ``BiFuse++ w/ SSIM'': our framework trained with structural similarity index (SSIM). 2) ``BiFuse++ w/ SPL'': our framework trained with spherical photometric loss proposed by \cite{ouraccv}. 3) \revise{``BiFuse++ w/ CAPL''}: our framework trained with our proposed Contrast-Aware Photometric Loss. The quantitative and qualitative results are shown in Table~\ref{tab:self-panosuncg} and Figure~\ref{fig:qualitative-self-panosuncg}, respectively.

Compared with our previous work 360-SelfNet~\cite{ouraccv}, our framework ``BiFuse++ w/ CAPL'' quantitatively improves 360-SelfNet by $16\%$ in RMSE and $23\%$ in MAE, as shown in Table~\ref{tab:self-panosuncg}. Compared to UniFuse, we improved $13\%$ in RMSE and $26\%$ in MAE. Since the architecture of 360-SelfNet only adopts cubemap projection as input, the benefit of equirectangular projection is discarded, and thus 360-SelfNet introduces much noise to predicted depth maps, as shown in Figure~\ref{fig:qualitative-self-panosuncg}.
Although UniFuse can predict sharper depth maps than 360-SelfNet does, there are still obvious errors around object/wall boundaries. In contrast, BiFuse++ is capable of recovering more details of objects and has smaller errors around the object boundaries. Such an improvement comes from our fusion approach. During the encoding process, we pass the fused feature maps ($f_{equi}^\prime$ and $f_{cube}^\prime$ in Figure~\ref{fig:our-fusion}) to layers of encoders to preserve more details, and thus BiFuse++ achieves sharper depth predictions in the end.

Compared to the training results without CAPL (BiFuse++ w/ SPL), ``BiFuse++ w/ CAPL'' improves the RMSE and MAE by $8\%$ and $13\%$, respectively (see Table~\ref{tab:self-panosuncg}). Such an improvement validates the reason why we design CAPL to prevent the network from focusing on low texture areas, i.e., to balance the difference between high-texture and low-texture areas. Comparing to training with SSIM loss (BiFuse++ w/ SSIM), the final performance is worse than training with spherical photometric loss. We have tried to apply other structure-based loss functions like weighted local contrast normalization (WLCN) proposed by \cite{activestereonet}, but the training fails to converge, and no reasonable related depth maps can be generated. Both SSIM and WLCN loss functions apply normalization to image patches, and we find such an operation eventually harms the training stability. In contrast, our Contrast-Aware Photometric Loss uses the standard deviation of image patches as the weighting of photometric loss instead of directly applying normalization, and thus we can prevent numerical instability. \newline

\begin{figure}
    \centering
    \includegraphics[width=\columnwidth]{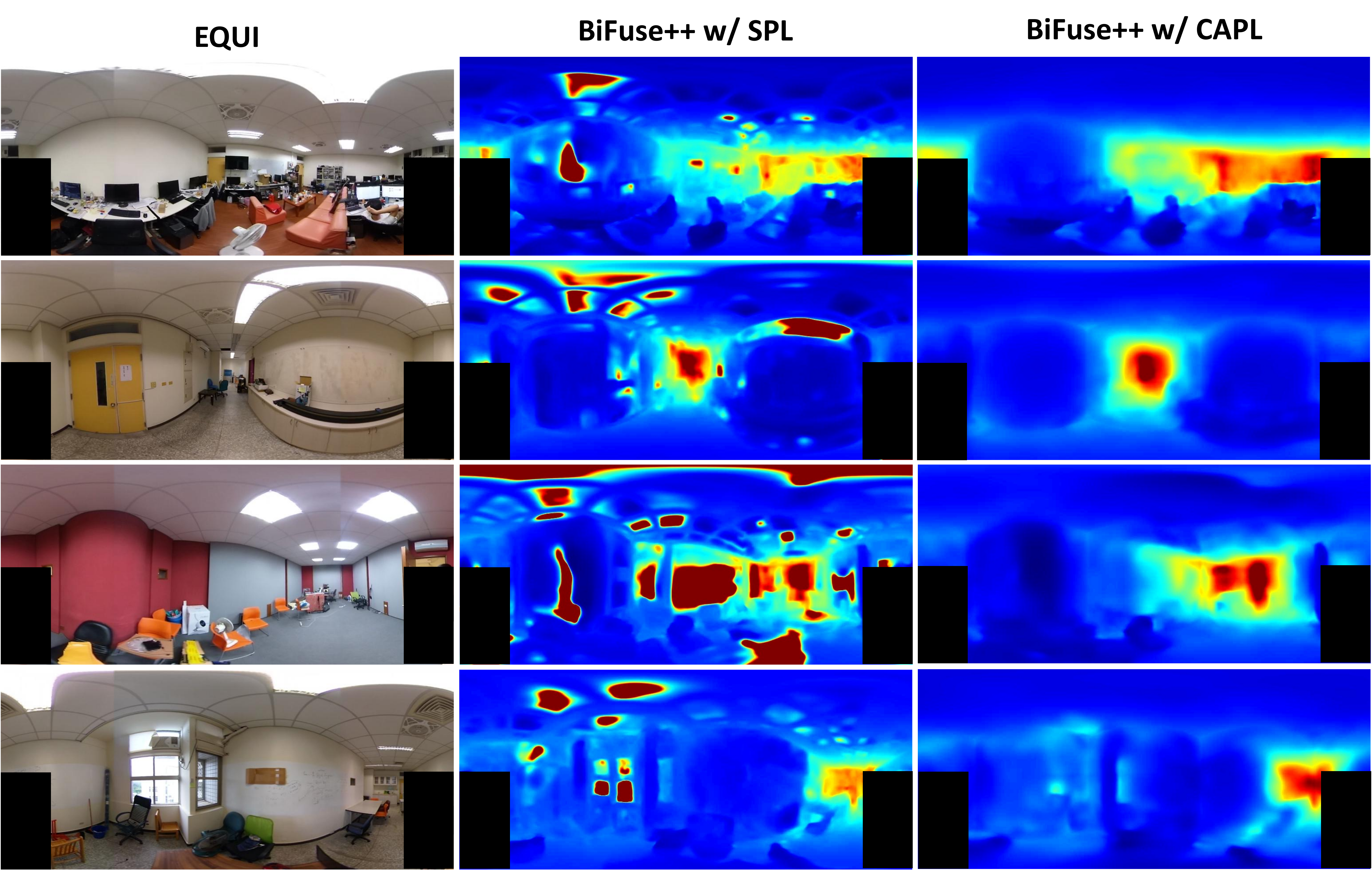}
    \caption{The effect of Contrast-Aware Photometric Loss (CAPL). The Spherical Photometric Loss (SPL) cannot deal with the low-texture area and thus produces unstable depth maps (red indicates a large depth value). Note that we mask out the photographer at the bottom left and right region.}
    \label{fig:ablation-loss}
\end{figure}

\noindent \textbf{Training on real-world videos.~}
To train BiFuse++ in real-world scenarios, we use the dataset we collected in 360-SelfNet~\cite{ouraccv}, in which there are 25 video sequences of 5 rooms recorded with RICOH THETA V, to apply the self-training strategy adopted in Section~\ref{sec:results-selfsupervised} for experiments. 
To ensure the training stability and the small baseline between consecutive frames, we extract the raw videos with 5 frames per second. Since we do not have laser scanners like LiDARs to collect the depth ground truths, we show the training results qualitatively. As addressed in Section~\ref{sec:ours}, directly applying spherical photometric loss \cite{ouraccv} to the real-world videos results in unstable depth prediction under low-texture areas such as walls or floor, since the corresponding photometric loss is ambiguous, i.e., there is a degeneration problem in the spherical photometric loss. Thus, we propose CAPL to prevent the network from focusing on these areas overly and we compare the training results before/after applying CAPL in Figure~\ref{fig:ablation-loss}. Without CAPL, the depth maps are noisy, especially on the wall and floor, while the results are smooth and stable after applying CAPL. Hence, our proposed BiFuse++ along with CAPL is a general self-supervised 360$^\circ$ depth estimation framework capable of estimating high-quality depth maps in both virtual and real-world environments. \newline

\begin{figure}
    \centering
    \includegraphics[width=\columnwidth]{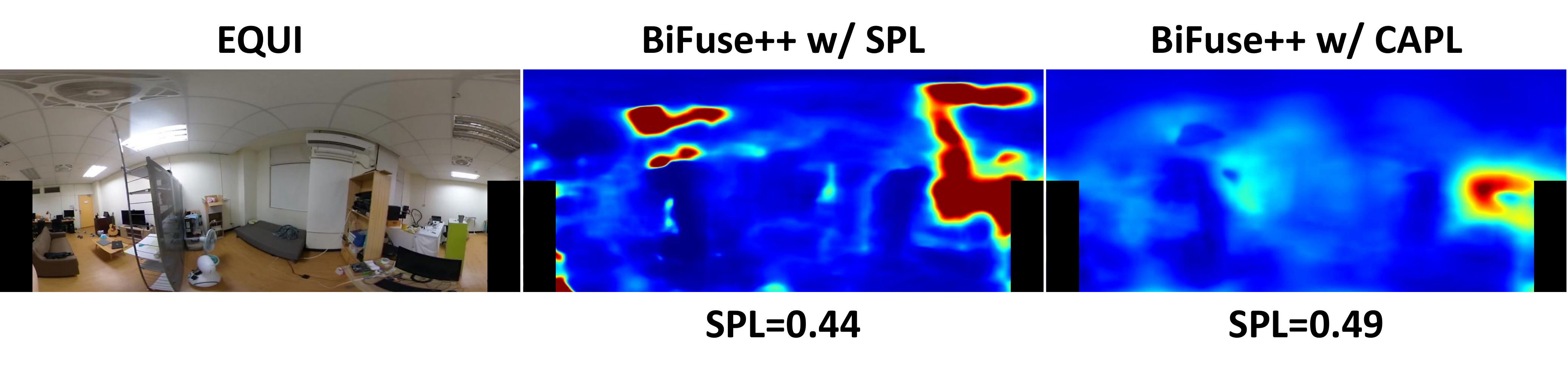}
    \caption{The Spherical Photometric Loss (SPL) is a degenerated loss. The BiFuse++ w/ SPL indeed reaches a lower SPL (0.44) compared to BiFuse++ w/ CAPL (0.49). However, the quality of BiFuse++ w/ SPL is worse than BiFuse++  w/ CAPL.}
    \label{fig:spl}
\end{figure}

\noindent \textbf{Spherical Photometric Loss Degeneration.~}
To investigate the reason why our CAPL can improve the results on real-world videos, we monitor the spherical photometric loss (SPL) value of ``BiFuse++ w/ SPL'' and ``BiFuse++ w/ CAPL''. For BiFuse++ w/ SPL, the average spherical photometric loss value over the validation set is 0.45, while the average loss value becomes 0.48 after applying CAPL. This indicates that the network with a lower average spherical photometric loss cannot always produce better depth maps. As the example shown in Figure~\ref{fig:spl}, the depth map with higher spherical photometric loss is better than the lower one. When there is no constraint on low-texture areas (w/ SPL), we find that the networks still tend to keep minimizing the spherical photometric error on these areas even if the corresponding value is already small. Since there is small intensity noise introduced by camera sensors between videos frames, the spherical photometric loss is impossible to be zero when the perfect depth prediction and camera motion are given. The minimizing behavior of networks in low-texture areas severely harms the training stability. Thus, our CAPL uses the standard deviation of neighboring pixels to directly enforce our networks not to overly focus on these areas because the standard deviation of low-texture areas is always small.

\section{Conclusion}
We propose ``BiFuse++'', the first bi-projection architecture for both self-supervised and supervised 360$^\circ$ depth estimation, extending our previous works 360-SelfNet~\cite{ouraccv} and BiFuse~\cite{bifuse}. To improve the efficiency and scalability, we propose a new fusion module that adopts a residual connection and removes the mask module from the original fusion module. In addition, we follow \cite{unifuse} that removes a redundant decoder and adopts a pixelshffule upsampling strategy to improve efficiency. We conduct experiments on three benchmark datasets of 360$^\circ$ depth estimation and achieve state-of-the-art performance in self-supervised scenarios and comparable performance with state-of-the-art approaches in supervised scenarios.

\noindent \textbf{Acknowledgments.}
This paper is the extended version of our previous works, 360-SelfNet~\cite{ouraccv} and BiFuse~\cite{bifuse}, and our project is funded by Ministry of Science and Technology of Taiwan (MOST110-2634-F-007-016, MOST110-2634-F-002-051, MOST111-2628-E-A49-018-MY4, and MOST111-2636-E-A49-003).

\ifCLASSOPTIONcaptionsoff
  \newpage
\fi

\bibliographystyle{IEEEtran}

\begin{IEEEbiography}[{\includegraphics[width=1in,clip,keepaspectratio]{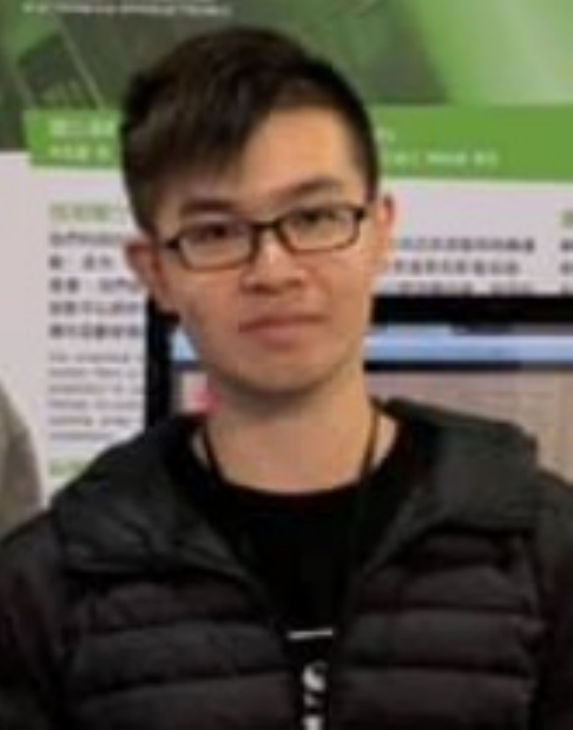}}]{Fu-En Wang}
    is~a Ph.D. student in Electrical Engineering from National Tsing Hua University (NTHU). He was a member of ASUS Ph.D. program in 2019, and became a research intern of Microsoft in 2021. His research interests include topics of computer vision such as scene understanding and 3D geometry. His research publications primarily focus on 360$^\circ$ perception including 360$^\circ$ depth estimation and 360$^\circ$ indoor layout estimation. In addition, he has collected several 360$^\circ$ benchmark datasets to inspire more researchers to work on 360$^\circ$ perception and tackle the issues of spherical projection.
\end{IEEEbiography}

\begin{IEEEbiography}[{\includegraphics[width=1in,clip,keepaspectratio]{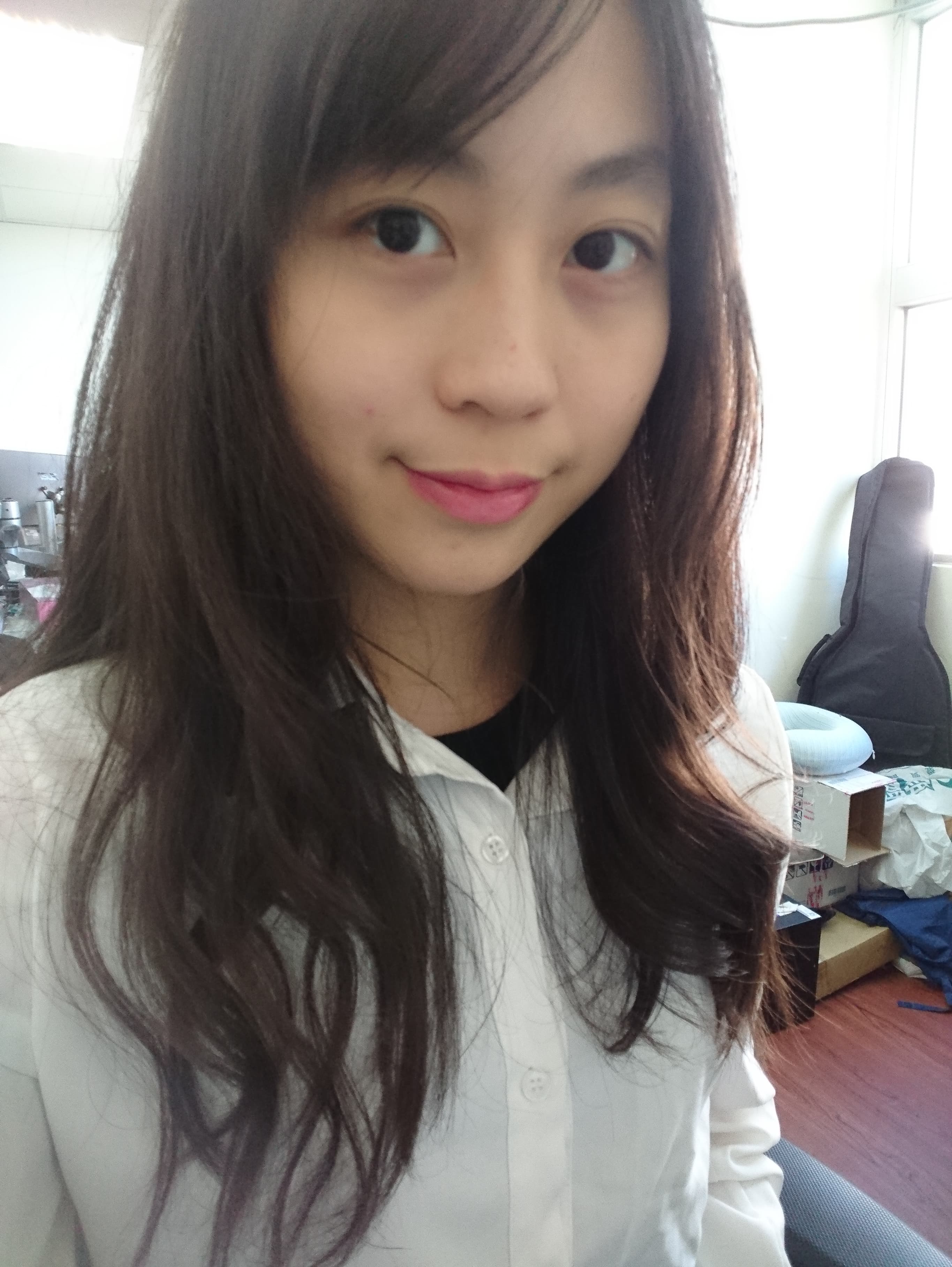}}]{Yu-Hsuan Yeh} 
    will~be a student studying master of science in computer vision (MSCV) at Carnegie Mellon University (CMU). Now she is an intern at Lumachain, working on vision-related projects. In 2021, she received her first MS degree in Artificial Intelligence Graduate Program at National Yang Ming Chiao Tung University (NYCU). Her research interests focus on Computer Vision and Deep Learning, including topics on scene understanding in the real-world domain, 360 images applications, and designing new panorama representations. Moreover, she was a co-first author of two CVPR papers related to 360 depth and layout tasks.
\end{IEEEbiography}

\begin{IEEEbiography}[{\includegraphics[width=1in,clip,keepaspectratio]{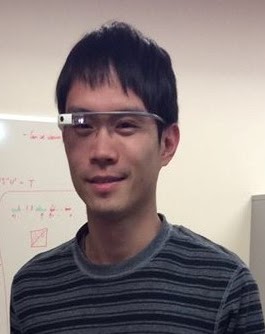}}]{Yi-Hsuan Tsai}
    is the Director of AI at Phiar Technologies, leading the AI team to improve real-world AR navigation. He was a senior researcher at NEC Laboratories America, working on fundamental computer vision and deep learning research. He received his PhD with Prof. Ming-Hsuan Yang at University of California, Merced, honored with the Graduate Dean's Dissertation Fellowship. Prior to that, he received his MS at University of Michigan, Ann Arbor and BS at National Chiao Tung University, Taiwan. He is the recipient of the Best Student Paper Honorable Mention award for ACCV'18. He has various research interests in computer vision and machine learning, with a focus on scene understanding, video analysis, domain adaptation, multi-modal learning, and representation learning.
\end{IEEEbiography}

\begin{IEEEbiography}[{\includegraphics[width=1in,clip,keepaspectratio]{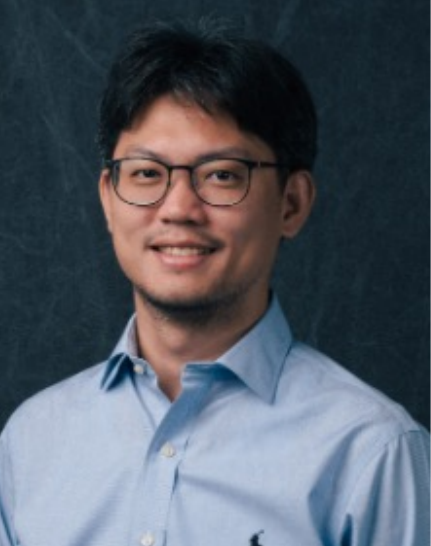}}]{Wei-Chen Chiu}
    received~the B.S. degree in Electrical Engineering and Computer Science and the M.S. degree in Computer Science from National Chiao Tung University (Hsinchu, Taiwan) in 2008 and 2009 respectively. He further received Doctor of Engineering Science (Dr.-Ing.) from Max Planck Institute for Informatics (Saarbrucken, Germany) in 2016. He joined Department of Computer Science, National Chiao Tung University as an Assistant Professor from August 2017 and got promoted to Associate Professor in July 2020. From May 2021, he was also hired as a Joint Appointment Research Fellow by the Mechanical and Mechatronics Systems Lab of  Industrial Technology Research Institute, Taiwan. His current research interests generally include computer vision, machine learning, and deep learning, with special focuses on generative models, multi-modal preception, and 3D recognition.
\end{IEEEbiography}

\begin{IEEEbiography}[{\includegraphics[width=1in,height=1.25in,clip,keepaspectratio]{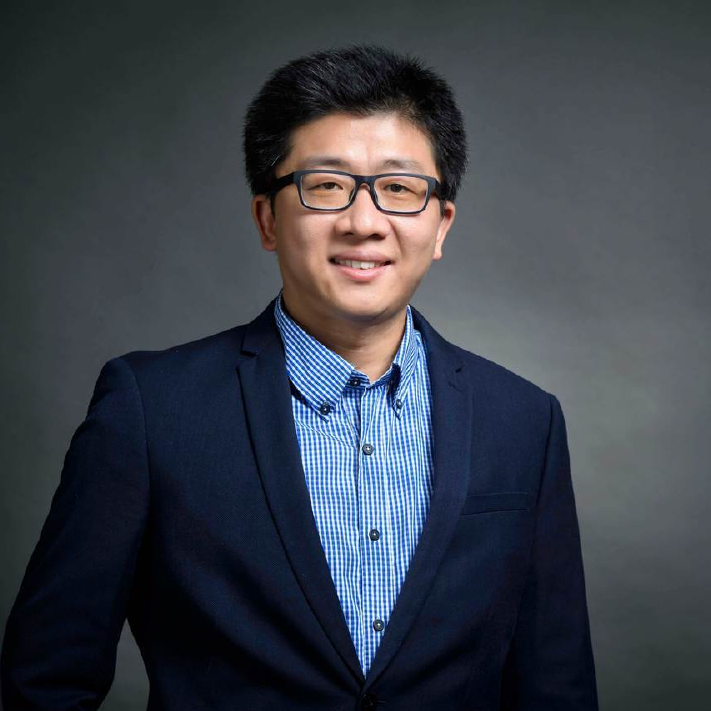}}]{Min Sun}
    is~an Associate Professor of the EE Department at National Tsing Hua University (NTHU) and the chief AI scientist at Appier, Inc. Before joining NTHU, he was a postdoc in CSE at UW. He graduated from UofM at Ann Arbor with a Ph.D. degree and Stanford University with an M.Sc. degree. His research interests include 2D+3D object recognition, 2D+3D scene understanding, and human pose estimation. He recently focuses on developing self-evolved neural network architectures for visual perception and designing specialized perception methods for 360-degree images/videos. He has won the best paper award in 3DRR 2007, was a recipient of the W. Michael Blumenthal Family Fund Fellowship in 2007, the Outstanding Research Award from MOST Taiwan in 2018, and Ta-You Wu Memorial Award From MOST Taiwan in 2018.
\end{IEEEbiography}

\end{document}